\documentclass[journal]{IEEEtran}   

\usepackage[pdftex]{graphicx}
\DeclareGraphicsExtensions{.pdf,.jpeg,.png}
\usepackage{amsmath}
\usepackage{array}
\usepackage[caption=false,font=footnotesize]{subfig}

\usepackage{gensymb}
\usepackage{cclicenses}
\usepackage{amsfonts}
\usepackage[keeplastbox]{flushend}
\usepackage{multirow}

\newcommand{\graspit}{\textit{GraspIt!}}

\newcolumntype{C}[1]{>{\centering\arraybackslash}m{#1}}
\usepackage{tikz}
\usepackage{color}
\definecolor{darkgreen}{RGB}{0,127,0}
\definecolor{darkblue}{RGB}{0,0,175}

\newcommand{\remindtext}[2]{{{#2}}}
\newcommand{\addedtext}[2]{{{#2}}}
\newcommand{\modifiedtext}[2]{{{#2}}}
\newcommand{\deletedtext}[2]{{}}
\newcommand{\secondrevisiontext}[2]{{{#2}}}

\begin{document}

\title{A Continuous Teleoperation Subspace with Empirical and Algorithmic Mapping Algorithms for Non-Anthropomorphic Hands}

\author{Cassie Meeker~\IEEEmembership{Student Member,~IEEE,} 
        Maximilian Haas-Heger~\IEEEmembership{Student Member,~IEEE,} \\ 
	and~Matei Ciocarlie~\IEEEmembership{Member,~IEEE}%
\thanks{This work was supported in part by the 
ONR Young Investigator Program award N00014-16-1-2026, and the NSF Career grant IIS-1551631.}%
\thanks{C. Meeker, M. Haas-Heger and M. Ciocarlie are with the Department of Mechanical Engineering, Columbia University, New York, NY, 10027 USA. (email: \{c.meeker, m.haas, matei.ciocarlie\}@columbia.edu)}%
\thanks{Manuscript received October 2020.}
}

\markboth{Transactions on Automation Science and Engineering}%
{Meeker \MakeLowercase{\textit{et al.}}: A Continuous Teleoperation Subspace with Empirical and Algorithmic Mapping Algorithms for Non-Anthropomorphic Hands}

\maketitle

\begin{abstract}
Teleoperation is a valuable tool for robotic manipulators in highly unstructured 
environments. However, finding an intuitive mapping between a human hand and 
a non-anthropomorphic robot hand can be difficult, due to the hands' 
dissimilar kinematics. 
In this paper, we seek to create a mapping between the human hand and a fully 
actuated, non-anthropomorphic robot hand that is intuitive enough to enable 
effective real-time teleoperation, even for novice users. To accomplish this, 
we propose a low-dimensional teleoperation subspace which can 
be used as an intermediary for mapping between hand pose spaces. 
We present two different methods to define the teleoperation subspace: an 
empirical definition, which requires a person to 
define hand motions in an intuitive, hand-specific way, and an 
algorithmic definition, which is kinematically independent, and uses objects 
to define the subspace. We use each of these definitions to create a 
teleoperation mapping for different hands. 
\addedtext{4-10}{One of the main contributions of this paper is the validation of both the empirical and algorithmic mappings with teleoperation 
experiments} controlled by \addedtext{2-4}{ten} novices and performed on two kinematically distinct 
hands. The experiments show that the proposed subspace is 
relevant to teleoperation, intuitive enough to enable control by novices, and 
can generalize to non-anthropomorphic hands with different kinematics.

\end{abstract}

\def\abstractname{Note to Practitioners}
\begin{abstract}
As robots move into our warehouses, workplaces, and homes, it is important to develop robotic controls  which are intuitive and easy for novices to use. In particular, teleoperation can be valuable to guide robots when they encounter situations which autonomous programs are not prepared to deal with. In this paper, we focus specifically on robotic grasping using non-anthropomorphic hands. Our method is intended for novice users to intuitively teleoperate such robots. We show that the teleoperation subspace we use can effectively enable pick-and-place tasks and in-hand manipulation tasks and that it is intuitive for novice operators. Our subspace outperforms state-of-the-art methods for pick-and-place tasks and performs as well as state-of-the-art methods for in-hand manipulation.

\end{abstract}



\section{Introduction}\label{sec:introduction}

\begin{figure}[t]
\centering
\vspace{0mm}
\begin{tabular}{r}
    \hspace{-4mm}
    \subfloat[\label{fig:side_by_side_manipulation}]{%
       \includegraphics[trim=0cm 1cm 0cm 0cm, clip, width=0.9\linewidth]{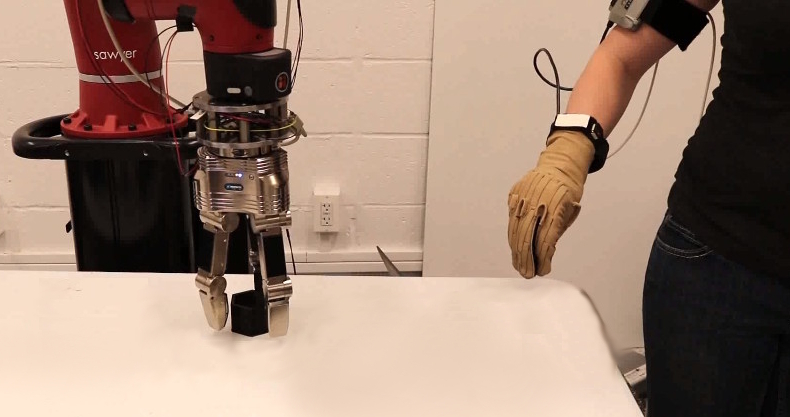}
    }
    \\ [-0.5mm]
        \hspace{-7mm}
     \subfloat[\label{fig:subspace_empirical_paradigm}]{%
       \includegraphics[trim=0cm 1cm 0cm 1cm, clip, width=0.9\linewidth]{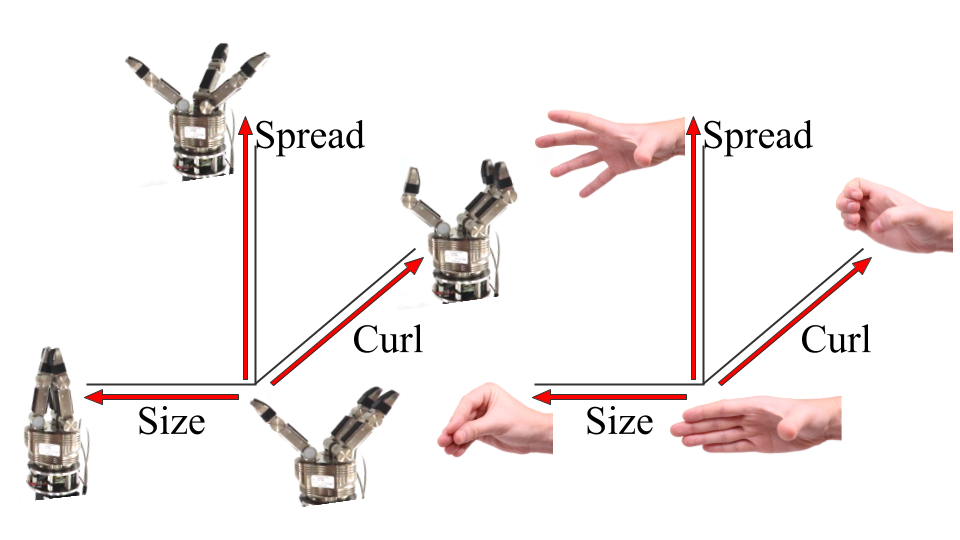} 
    } 
\end{tabular}
\caption{\textbf{(a)} Teleoperated manipulation where the operator's hand movements are recorded and used to drive a non-anthropomorphic robotic hand. To achieve this, a mapping between the human and robot hand kinematics is required. \textbf{(b)} A teleoperation subspace used as an intermediary between the pose spaces of different hands. Here we show how the motions that are associated with the subspace basis vectors can be intuitively defined by a user, based on the hand's kinematics.}
\vspace{-2mm}
\end{figure}

Teleoperation is a valuable tool for robotic manipulators in
unstructured environments, where a wide array of scenarios and objects
can be encountered. In such conditions, the robot can rely on human
cognition to deal with corner cases faster and more easily than
autonomous manipulation planners. 

\addedtext{1-1}{As robots and teleoperation become more common in our everyday lives, from our homes~\cite{doelling2014} to our workplaces~\cite{savela2018}, it becomes increasingly likely that the people who will need to work with robots will not be robotic experts, but novices.} Therefore, an important research direction for 
robot teleoperation aims to make the controls available to the operator 
as intuitive as possible: intuitive controls minimize the training time 
required for human teleoperators and can make teleoperation more 
accessible to novices. They also ensure a safe and effective workflow. 

For manipulation, teleoperation controls which harvest the user's 
hand motions, rather than using a joystick or a point-and-click interface, 
can provide an intuitive and user friendly interface~\cite{ferre2007}, 
because they harness motions which are already natural to the teleoperator. 
An example of this workflow is shown in Figure~\ref{fig:side_by_side_manipulation}.

Teleoperating a robot hand using a human hand as input requires a 
teleoperation mapping, which tells the robot hand how to move in 
response to movements of the human hand. Robot hand designs that 
are fully-actuated and highly anthropomorphic allow for a direct 
joint mapping to the human hand and thus are intuitive for a human 
to teleoperate; however, the hardware tends to be fragile
and expensive. In contrast, non-anthropomorphic hands have proven to be 
robust and versatile in unstructured environments. However, finding an easy
or intuitive mapping between the human hand and a non-anthropomorphic 
robot hand can be difficult, due to the different joint configurations,
different axes, different numbers of fingers, or any number of
dissimilarities between the hands.

In this paper, we seek to create a mapping between the human hand
and a fully actuated but non-anthropomorphic robot hand that is intuitive enough to enable
effective real-time teleoperation, even for novice users.

We propose a subspace relevant to teleoperation. This subspace is 
an intermediary which allows us to map between the pose spaces of different hands. 
By projecting the pose of the master hand 
into the teleoperation subspace, which it shares with the slave 
hand, and then projecting from the teleoperation subspace into the pose
space of the slave hand, we can enable real-time teleoperation.

At a conceptual level, each of the basis vectors that define the
subspace corresponds to a hand motion: hand opening, finger curl, and
finger spread (Figure~\ref{fig:subspace_empirical_paradigm}). While
these concepts are natural for the human hand, we need to also define
them in the context of non-anthropomorphic robot hands. We show that
this process can be done empirically: in this formulation, the person
creating the teleoperation mapping defines what the motions of `open',
`curl', and `spread' mean for a specific hand. In this way, the
mapping is tied to hand kinematics, since the hand motions mean
different things for different hands. 

One shortcoming of using an empirical mapping is its reliance on human
intuition: effective teleoperation could be attributed to either the
structure of the subspace, or to the hand-specific intuition
provided by the person creating the mapping.  

We therefore propose a second, algorithmic, method where we formalize
the notion of the hand motions used to define the subspace. Rather
than considering, for example `hand opening' as an intuitive concept
to be defined by the person creating the mapping, this paradigm
considers hand opening as the hand grasping a series of incrementally
larger objects. In this way, we can use a set of objects to provide
the same understanding of hand motions as the user provided in the
empirical method.

This definition of the subspace is done exclusively through an object
set, and is kinematically-independent. However, it lends itself to the
algorithmic creation of a teleoperation mapping for any hand. We
introduce a method which uses this algorithmic definition to generate
subspace mappings for hands in a fully automated fashion
(Section~\ref{sec:algorithmic_mapping}). We aim to show that this
algorithmic mapping also enables effective teleoperation for novices,
implying that the value of the teleoperation subspace does not derive
exclusively from hand-specific human intuition used to create it.

We create teleoperation mappings for two non-anthropomorphic robotic
hands using both the empirical mapping and the algorithmic mapping. In
teleoperation experiments with \addedtext{2-4}{ten} novices, we show that the mappings
created with both these paradigms can enable teleoperation as fast as
or faster than state-of-the-art teleoperation methods. Overall, the main
contributions of this paper are:
\begin{itemize}
\item We introduce a teleoperation
  subspace as an intuitive way to map between human and robotic hand poses. 
\item We provide an empirical method to define the subspace and to
  create a projection into the subspace. 
\item We provide an algorithmic method for defining the subspace that is independent
  of hand kinematics.  We are the first to show that an
  automated method for generating a teleoperation
  mapping can enable online teleoperation which is intuitive for
  novices.
\item We validate the 
  teleoperation subspace mappings on \remindtext{2-4}{ten novice users} and two 
  different robotic hands.
\end{itemize}

In an earlier version of this study~\cite{meeker2018}, we
introduced the concept of a teleoperation subspace defined exclusively
via empirical mapping, and validated it with teleoperation experiments
on a single robotic hand for a single task. Here, we show that the
subspace can be defined in a hand-independent fashion by considering
variations in the grasped object shape, and introduce an
automated process for creating mappings into this subspace. 
\addedtext{4-10}{We also expand our validation of the subspace considerably with experiments that test our mapping methods with two
manipulation tasks on two kinematically distinct robotic hands. These experiments show that the subspace is relevant to teleoperation for multiple manipulation tasks and it is more intuitive for novices than the state-of-the-art.}.

\section{Related Work} \label{sec:related_work}

The most common teleoperation mappings are: joint 
mapping~\cite{cerulo2017}, fingertip mapping~\cite{rohling1993}, and pose 
mapping~\cite{pao1989}. 

Joint mapping imposes the values of the human joint angles directly onto the robot joints with little or no transformation~\cite{cerulo2017}. This is most useful if the slave hand has similar kinematics to the master hand~\cite{liarokapis2013}. 

Fingertip mapping is the most common 
teleoperation mapping method. Cartesian positions of the human fingertips (found with forward kinematics) are the control input. After scaling, these
positions represent the desired robot fingertip positions. 
Joint angles which allow the robot to achieve this desired position are
found with inverse kinematics.

For fingertip and joint mapping, how to reconcile kinematic differences 
between the human and robotic workspaces is an open question. To solve this problem, researchers combine 
different types of mappings~\cite{colasanto2013, chattaraj2014}, use virtual object mapping ~\cite{griffin2000}, 
optimize distances in task space~\cite{liarokapis2013}, use error 
compensation~\cite{rohling1993}, and alter the robotic hand frame to minimize the workspace differences~\cite{geng2011}. 

Pose mapping attempts to interpret the function of the human grasp, rather than replicate literal hand position. In this type of control, the robot is intended to replicate the pose of the human hand. Sometimes, this requires identifying the human pose before mapping between the human and the robot~\cite{ekvall2004, wojtara2004}. Others perform
pose mapping in an end-to-end fashion, either with transformation matrices~\cite{pao1989} or, more recently, neural networks~\cite{li2019}. Li et al. tested their method with real time experiments, but they use vision as input.  We would like to avoid vision-based methods because they 
require environments which are well-lit and have few occlusions, conditions that cannot be guaranteed in unstructured environments. Additionally, even in the 
end-to-end pose mappings, outside of a discrete set of known poses, pose mapping can lead to unpredictable hand motions. Because of this, pose mapping is less common than either fingertip or joint mapping.

In this paper, we use a low-dimensional mapping to define grasping.
Another low dimensional grasping space was found by Santello et al.~\cite{santello1998}, though these `postural synergies' are specific to humans. The concept of postural synergies has been applied to robots by finding synergies of robot hands by finding robot poses that 
resemble grasping poses for human hands, and then performing \addedtext{3-6}{principal component analysis (PCA)} on 
those poses. The poses are either found through joint mapping~\cite{ficuciello2013}, 
pose mapping~\cite{kim2016}, or human intuition~\cite{ficuciello2012}

Other works use low dimensional latent variables not based on 
synergies to approximate human poses in non-anthropomorphic models. 
Gaussian process latent 
variable models (GP-LVM) can enable teleoperation of humanoid robots. In 
some formulations, the latent space changes with every different master-slave 
pairing~\cite{shon2006}. In other formulations, multiple robots and a 
human share the same latent space~\cite{delhaisse2017}.

Training data driven mappings, like some pose mappings~\cite{ficuciello2012}, 
or GP-LVMs, requires the user to create many corresponding poses between the 
human and robotic hands. Creating these poses is tedious 
and time consuming. 

There are works which, like our algorithmic mapping, try to create 
teleoperation mappings without requiring that the user provide intuition about 
hand kinematics. 

Kheddar et al. proposed high level abstraction teleoperation, where 
the operator manipulates a virtual environment and a bilateral transform 
translates changes in the virtual environment into commands for the robot
~\cite{kheddar1997}. 
The gripper control is object based, i.e. the robot must manipulate and 
transport an object in the real world in the same way it is being manipulated 
in the virtual environment~\cite{kheddar2001}. Although they describe several 
possible ways to transform between the human and robotic hands, their ultimate 
solution is autonomous. 
Learning high level tasks from demonstrations is reviewed elsewhere~\cite{chernova2014}.

Kang et al. also introduced an object based approach to identifying human 
grasps using the contact web~\cite{kang1997}. Once the human grasp has been 
identified, the robot hand is shaped based on virtual fingers and the human 
grasp. However, the user still provides an understanding of how the robot 
functions - for each new robot, they assign the fingers as being a primary 
finger, a secondary finger, or a palm. 

Finally, Gioioso et al. defined an object based approach for mapping between 
hands with dissimilar kinematics using virtual 
objects~\cite{
gioioso2013mapping, salvietti2014, salvietti2013, salvietti2016}. 
This work replicates the deformation of the virtual object in the human hand with 
the virtual object in the robot hand. This is the first time a (virtual) object 
set was used to define a teleoperation mapping. However, the 
authors have reported varying performance for the same hand with different 
number of virtual points and different numbers of synergies, meaning that 
creating the mapping for each hand requires the user to tune control parameters.
Of this body of work, only two publications show 
the method to be feasible for robotic hand manipulation with online 
teleoperation experiments~\cite{salvietti2013, salvietti2016}. The first work only considers spherical virtual objects~\cite{salvietti2013}. The method was later extended to a virtual object of any shape, but does not consider scenarios where the slave has fewer contact points than the master, and it assumes the the movement of the master contacts can be represented by a homogeneous transform~\cite{salvietti2016}. In our experiments, neither of these assumptions hold true.

As a final note about the pose mapping and synergy based methods, to our knowledge, only three publications in the literature demonstrate a pose or synergy mapping to be feasible for hand manipulation with online teleoperation experiments~\cite{salvietti2016, salvietti2013, li2019}. None of these were shown to be effective for a non-anthropomorphic hand, which is the focus
of this study. 

\modifiedtext{4-4}{Though there has been a significant effort in the literature to map between
human and robotic hands, and to use low dimensional spaces in order to achieve
this mapping}, most works validate their proposed 
teleoperation methods on one or two expert users or perform their experiments 
in simulation (e.g.~\cite{cerulo2017, rohling1993, pao1989, chattaraj2014, 
colasanto2013, griffin2000, ekvall2004, wojtara2004, 
gioioso2013mapping, salvietti2014}). 
We were only able to find two works that validate a proposed 
teleoperation method with novice users on a physical robot~\cite{li2019, salvietti2013}. 
Both of these works validated their method with five novices users teleoperating 
a single robotic hand. \modifiedtext{4-4}{To our knowledge, we are the first to validate with real-time experiments that the
subspace we propose is relevant to teleoperation for multiple robotic
hands, as well as the human hand.} In this paper, we validate our work over two different tasks 
with a total of ten novice users, using  
two different robotic hands.
Our experiments show that the mappings we propose are intuitive and encode 
information relevant to teleoperation.

\section{Teleoperation Subspace}

As a general concept, we posit that, for many hands, there exists a \modifiedtext{4-6}{three dimensional manifold $\boldsymbol T$} that can encapsulate the range of movement needed for teleoperation. The three dimensions of $\boldsymbol T$ correlate to certain hand motions: opening and closing the hand, spreading the fingers, and curling the fingers. We will refer to these as the size $\boldsymbol\sigma$, spread $\boldsymbol\alpha$, and curl $\boldsymbol\epsilon$ basis vectors, respectively.

We chose these bases on intuition, guided by Santello's research of postural synergies~\cite{santello1998}. Since Santello et al. used a linear dimension reduction method to find postural synergies, we assume the subspace described here is also linear. \remindtext{4-7}{We do not use the exact postural synergies found by Santello because we empirically found it easier to map to robotic hands (note that Santello's synergies are strictly human specific) if we `decoupled' Santello's two synergies into three. It is also easier to explain the control to novices with decoupled movements.} \secondrevisiontext{1-1}{We show evidence here that our selected basis vectors are appropriate for grasping and certain manipulation tasks, and believe they will generalize to others as well; however, this selection is ultimately arbitrary. The literature reports that humans can perform manipulation using varying numbers of synergies, depending on the task~\cite{todorov2004}, so different subspace dimensionalities can be explored and compared in future work.}

We assume that many hands will be able to project their pose spaces into $\boldsymbol T$. If this is possible, $\boldsymbol T$ is 
embedded as a subspace in the pose space of the hand. $\boldsymbol T$ is thus a subspace ``shared" by all hands that can project their pose space into $\boldsymbol T$. 

\begin{figure}[t]
\centering
\vspace{2mm}
\begin{tabular}{r}
\includegraphics[trim=5.37cm 6.0cm 2.1cm 4.5cm, clip, width=0.9\linewidth]{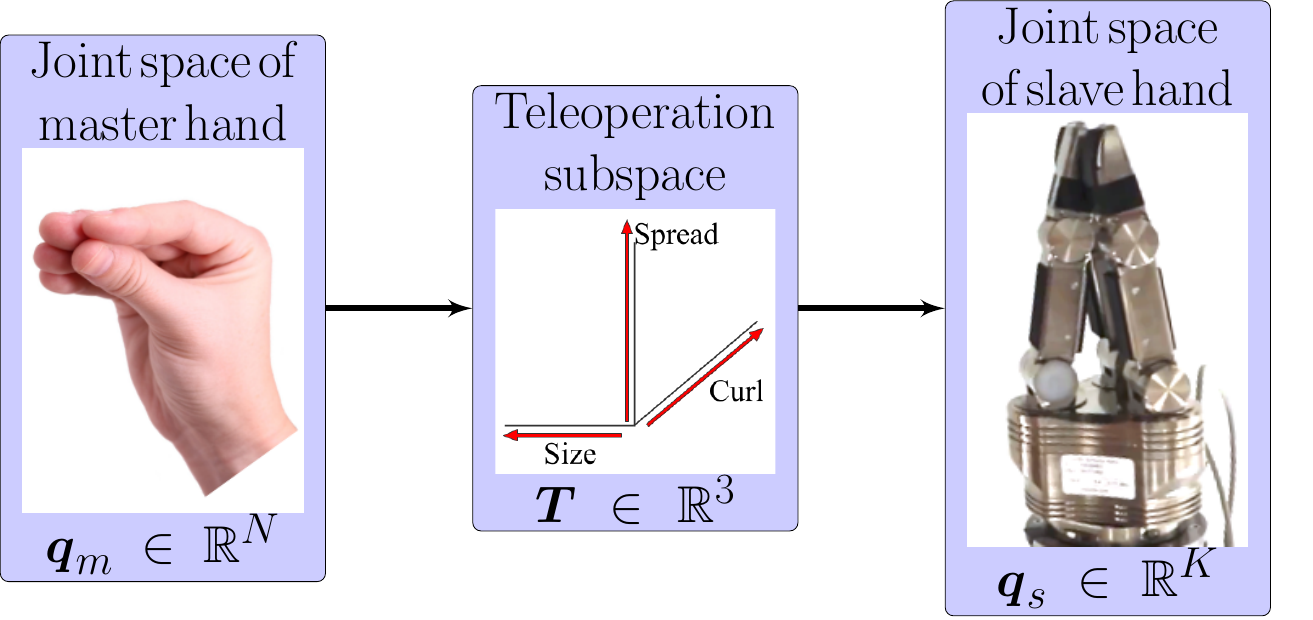}
\end{tabular}
\vspace{-2mm}
\caption{Steps to enable real time teleoperation using teleoperation subspace}
\label{fig:subspace_mapping_flow}
\vspace{-4mm}
\end{figure}

To teleoperate using $\boldsymbol T$, there are two steps (Figure ~\ref{fig:subspace_mapping_flow}):
\begin{enumerate}
\item Given joint values of the master hand, find the equivalent pose $\boldsymbol \psi$ in teleoperation subspace $\boldsymbol T$.
\item Given $\boldsymbol \psi$ computed above, find the joint values of the slave hand, and move the slave hand to these values.
\end{enumerate}

In order to enact the teleoperation steps, we must first define the mapping between $\boldsymbol T$ and the relevant pose spaces.

\subsection{Teleoperation Subspace Mapping}

For a given hand with $N$ joints, projecting from joint space $\boldsymbol q \in \mathbb{R}^N$ 
(we use pose space and joint space interchangeably) into 
teleoperation subspace $\boldsymbol T$ requires an origin pose $\boldsymbol o \in \mathbb{R}^{N}$, a 
projection matrix $\boldsymbol A \in \mathbb{R}^{N \times 3}$, and a scaling factor $\boldsymbol \delta \in \mathbb{R}^{3}$.
\subsubsection{Origin $\boldsymbol o$}
This hand-specific, ``neutral" origin 
pose $\boldsymbol o \in \mathbb{R}^N$ represents a hand position which will standardize the data as
we project between joint space and $\boldsymbol T$.
\begin{equation}
\boldsymbol o = [o_1, o_2, ... , o_N]
\end{equation}

The origin pose of 
the master is arbitrary; however, it is crucial that the origin 
pose of the slave corresponds to the master's origin. 
The two hands should assume approximately the same shape 
while positioned at their respective origins. 

\subsubsection{Projection Matrix $\boldsymbol A$}
The projection matrix $\boldsymbol A \in \mathbb{R}^{N \times 3}$ is hand
specific and consists of three basis vectors $\boldsymbol \alpha_H, \boldsymbol \sigma_H, \boldsymbol \epsilon_H \in \mathbb{R}^N$. 
Whereas $\boldsymbol \alpha, \boldsymbol \sigma$, and $\boldsymbol \epsilon$
represent the general concept of a hand motion, $\boldsymbol \alpha_H, \boldsymbol \sigma_H$, 
and $\boldsymbol \epsilon_H$ are the projection of that motion
into the pose space of hand $H$.
\\[-1em]
\begin{eqnarray}
\boldsymbol A = [\boldsymbol \alpha_H, \boldsymbol \sigma_H, \boldsymbol \epsilon_H] \\ 
\boldsymbol \alpha_H = [\alpha_{H 1}, \alpha_{H 2}, ... , \alpha_{H N}]^\top \\
\boldsymbol \sigma_H = [\sigma_{H 1}, \sigma_{H 2}, ... , \sigma_{H N}]^\top \\
\boldsymbol \epsilon_H = [\epsilon_{H 1}, \epsilon_{H 2}, ... , \epsilon_{H N}]^\top
\end{eqnarray}

\subsubsection{Scaling Factor $\boldsymbol \delta$}

We wish to normalize such that any configuration in pose space will project to 
a pose in $\boldsymbol T$ whose value is less than or equal to 1 along each of the basis 
vectors. We therefore require a scaling factor $\boldsymbol \delta \in \mathbb{R}^{3}$: 
\begin{equation}
\boldsymbol \delta = [\delta_\alpha, \delta_\sigma, \delta_\epsilon].
\end{equation}

To calculate $\boldsymbol \delta$, we evaluate poses which illustrate the extrema of the 
hand's kinematic limits along the basis vectors. 
Once we select the poses for a hand, we project them 
from pose space into $\boldsymbol T$ using $\boldsymbol\psi = (\boldsymbol q-\boldsymbol o) \cdot \boldsymbol A$, where $\boldsymbol \psi \in \boldsymbol T$. From this set of $\boldsymbol \psi$s, we find the minimum and maximum along $\boldsymbol\alpha$
$\alpha_{min}$ and $\alpha_{max}$, respectively. From this, we calculate 
$\delta_{\boldsymbol\alpha}$ as:
\begin{eqnarray}
\alpha_{range} = abs(\alpha_{max}) + abs(\alpha_{min}) \\
\delta_{\boldsymbol\alpha} = 
	  \begin{cases}
         0 & \text{if $\alpha_{range} = 0$ } \\
         1/\alpha_{range} & \text{otherwise.}
      \end{cases} \label{eq:delta}
\end{eqnarray}
Finding $\delta_{\boldsymbol\sigma}$ and $\delta_{\boldsymbol\epsilon}$ uses 
the same calculation.

To project 
from $\boldsymbol T$ back to pose space, we also require an inverse scaling factor $\boldsymbol \delta^*$:
\begin{eqnarray}
   \boldsymbol \delta^* = [\delta^*_\alpha, \delta^*_\sigma, \delta^*_\epsilon] \\
   \delta^*_\alpha = 
      \begin{cases}
         0 & \text{if $\delta_\alpha = 0$ } \\
         1/\delta_\alpha & \text{otherwise}
      \end{cases} \label{eq:delta_star}
\end{eqnarray}
where we find $\delta^*_\sigma$ and $\delta^*_\epsilon$ with similar calculations.

\secondrevisiontext{AE-1}{$\delta$ is the only mapping variable that changes for each human teleoperator. In all other cases, once the mapping is created, the user defined variables are constant.}

\subsubsection{A Complete Projection Algorithm}
To project between teleoperation subspace $\boldsymbol T$
and joint space $\boldsymbol q$, we use the hand-specific matrix $\boldsymbol A$, the
origin $\boldsymbol o$, and the scaling factor $\boldsymbol \delta$:
\begin{eqnarray}
\boldsymbol \psi = ((\boldsymbol q-\boldsymbol o) \cdot \boldsymbol A) \odot \boldsymbol\delta  \label{eq:to_subspace}\\
\boldsymbol q = ((\boldsymbol \psi \odot \boldsymbol \delta^*) \cdot \boldsymbol A^\top) + \boldsymbol o \label{eq:from_subspace}
\end{eqnarray}
where $\odot$ represents element-wise multiplication.

Eq.~\ref{eq:to_subspace} projects from 
the master hand's pose space into the shared
teleoperation subspace and then Eq.~\ref{eq:from_subspace}
projects from the teleoperation subspace
into the slave hand's pose space. 

So, given the joint angles of the master hand, we are able to
calculate the joint angles of the slave hand using:
\begin{eqnarray} \label{eq:complete_projection_algorithm}
\boldsymbol q_s = (((\boldsymbol q_m-\boldsymbol o_m) \cdot \boldsymbol A_m) \odot \boldsymbol \delta_m \odot \boldsymbol \delta^*_s) \cdot \boldsymbol A_s^\top + \boldsymbol o_s.
\end{eqnarray}

Now that we have formalized $\boldsymbol T$  and the variables required to map 
between pose space and the subspace, we propose two different methods
to define the mapping. The first, empirical method, is dependent on hand
kinematics and relies on the intuition of the person creating the mapping. 
The second, algorithmic mapping, is created automatically, using a definition
of the subspace which is independent of hand kinematics.

\section{Empirically Defining the Subspace Mapping}\label{sec:empirical_mapping}

\begin{table*}[t!]
\vspace{3mm}
\caption{Process to empirically define the projection matrix for the teleoperation subspace mapping}
\label{tab:empirical_projection_matrix_creation}
\vspace{-3mm}
\centering
\begin{tabular}{C{.13\linewidth} C{.27\linewidth}  C{.12\linewidth}   C{.35\linewidth} }
	Hand Motion  &  Motion Defined by Hand-Specific Kinematics  &  Joints which affect the motion  & Basis vector  \\ \hline

	Finger Spread & \includegraphics[width=0.9\linewidth]{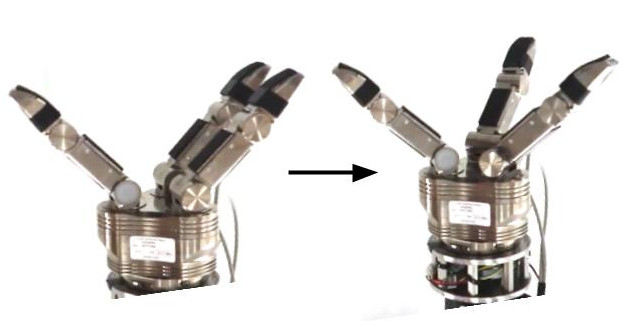} & \includegraphics[width=0.9\linewidth]{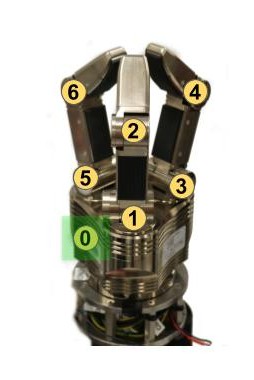}& $\boldsymbol\alpha_{schunk} = [1, 0, 0, 0, 0, 0, 0]^\top$ \\ [-1em]
Hand Opening & \includegraphics[width=0.9\linewidth]{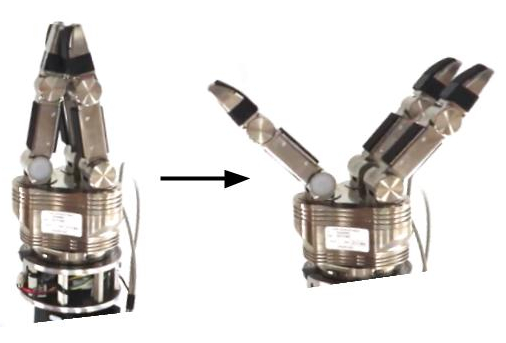} & \includegraphics[width=0.9\linewidth]{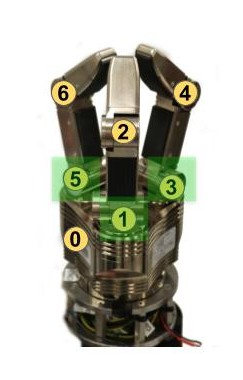}& $\boldsymbol \sigma_{schunk} = [0, 0.577, 0, 0.577, 0, 0.577, 0]^\top$ \\ [-1em]
Finger Curl & \includegraphics[width=0.9\linewidth]{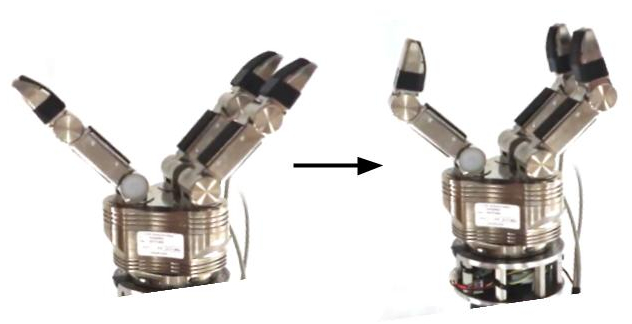} & \includegraphics[width=0.9\linewidth]{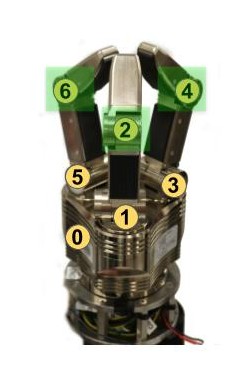}& $\boldsymbol\epsilon_{schunk} = [0, 0, 0.577, 0, 0.577, 0, 0.577]^\top$ \\ 
\end{tabular}
\vspace{-7mm}
\end{table*}

\begin{figure}[t]
\centering
\vspace{2mm}
\begin{tabular}{cccc}
\includegraphics[trim=7cm 2cm 7cm 8cm,clip, width=.2\linewidth]{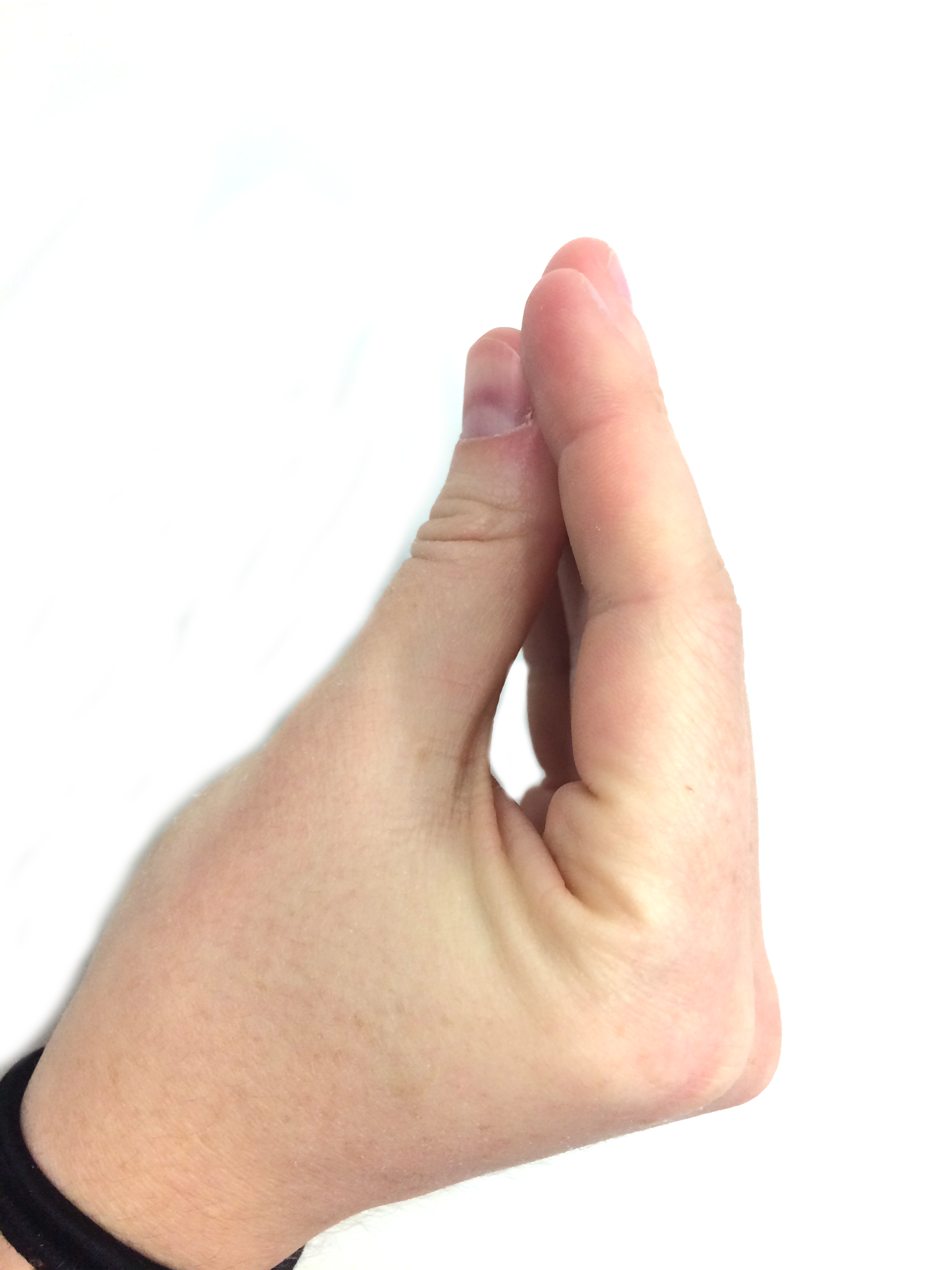}\hspace{.1cm}
\includegraphics[trim=0cm 0cm 0cm 0cm,clip, width=.2\linewidth]{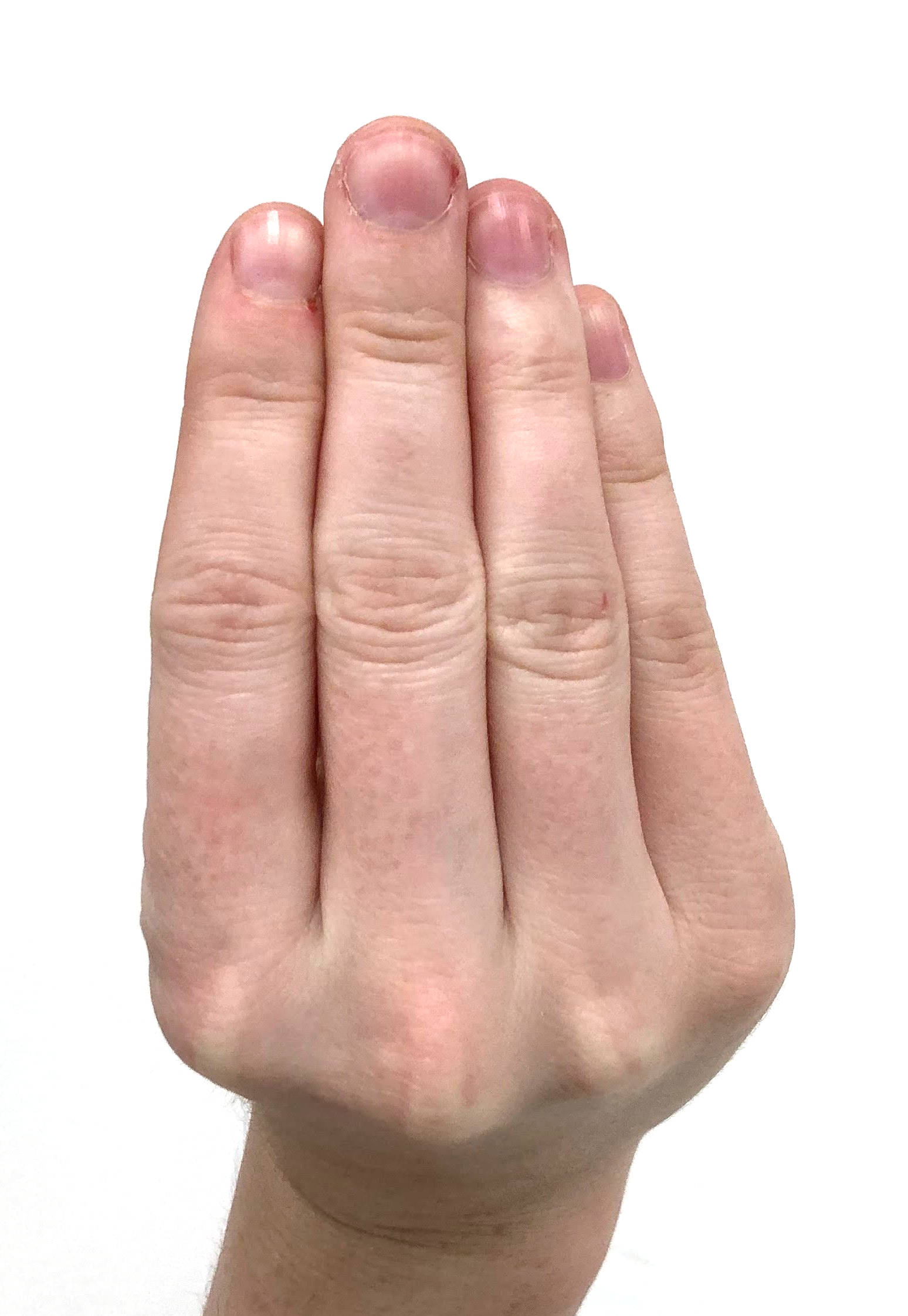}\hspace{.1cm}
\includegraphics[trim=0cm 5cm 0cm 2cm,clip, width=.2\linewidth]{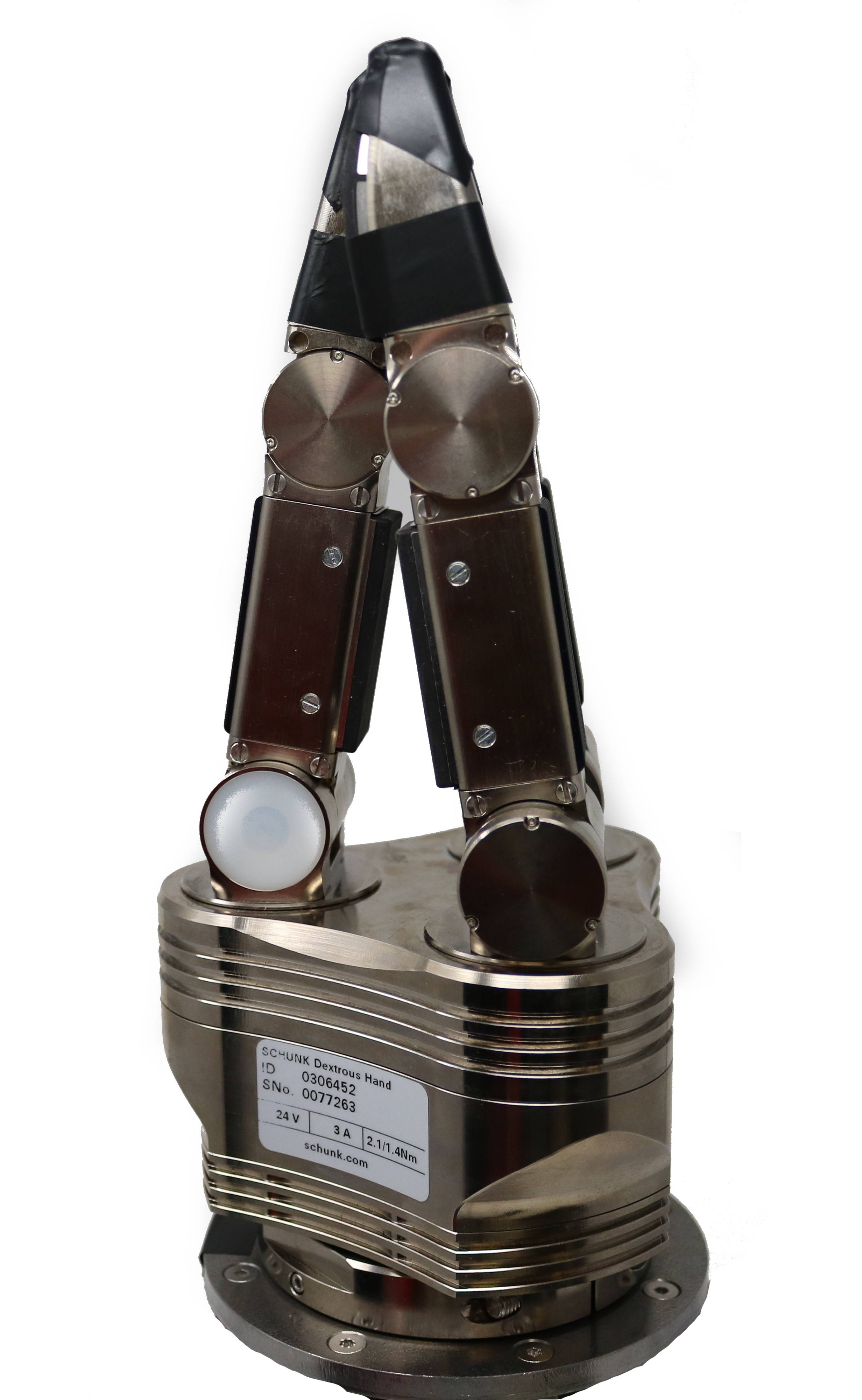}\hspace{.1cm}
\includegraphics[trim=40cm 20cm 31cm 32cm,clip, width=.13\linewidth]{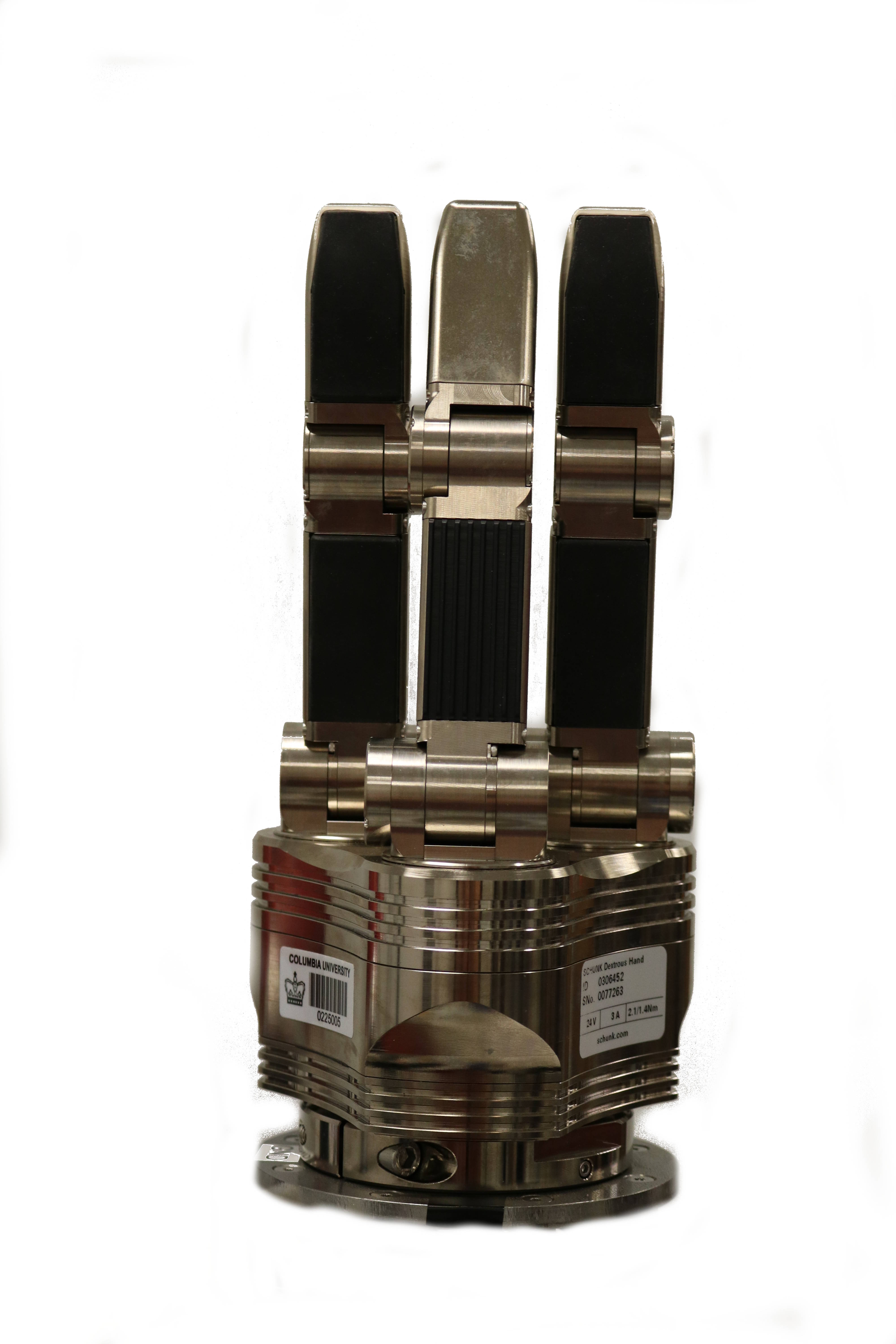}\\
\parbox[c]{.44\linewidth}{\footnotesize \centering Origin pose of the human hand.}\hspace{.1cm}
\parbox[c]{.44\linewidth}{\footnotesize \centering Origin pose of the Schunk SDH.}\hspace{.1cm}
\vspace{-2mm}
\end{tabular}
\caption{Origin poses of two example hands.}
\label{origin_pose}
\vspace{-4mm}
\end{figure}

\begin{figure}[t]
\centering
\setlength{\tabcolsep}{-1mm}
\begin{tabular}{cccc}
\includegraphics[trim=0.1cm 0.1cm 0cm 0.1cm,clip, width=.16\linewidth]{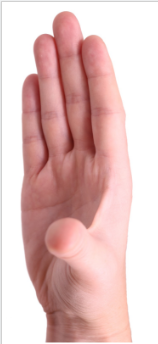} \hspace{1mm}
\includegraphics[trim=0cm 0cm 0cm 0cm,clip, width=.2\linewidth]{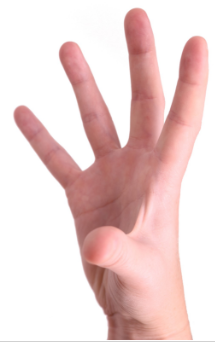} \hspace{1mm}
\includegraphics[trim=0cm 0cm 0cm 0cm,clip, width=.2\linewidth]{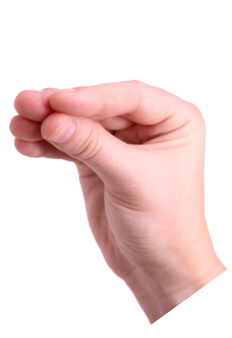} \hspace{1mm}
\includegraphics[trim=0cm 0cm 0cm 0cm,clip, width=.2\linewidth]{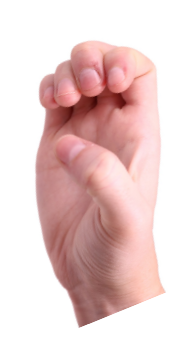}\\
\parbox[c]{.22\linewidth}{\footnotesize \centering Pose for maximum along $\boldsymbol\sigma_{human}$, minimum along $\boldsymbol\alpha_{human}$.}  \hspace{1mm}
\parbox[c]{.22\linewidth}{\footnotesize \centering Pose for maximum along $\boldsymbol\alpha_{human}$.}  \hspace{1mm}
\parbox{.22\linewidth}{\footnotesize \centering Pose for minimum along $\boldsymbol\epsilon_{human}$, and minimum along $\boldsymbol\sigma_{human}$.} \hspace{1mm}
\parbox{.22\linewidth}{\footnotesize \centering Pose for maximum along $\boldsymbol\epsilon_{human}$.} \\
\end{tabular}
\caption{ Poses which demonstrate the human hand's kinematics limits along the basis vectors of $\boldsymbol T$. To see the poses in the context of the teleoperation subspace, refer to Figure~\ref{fig:subspace_empirical_paradigm}.}
\label{calibration_poses}
\vspace{-4mm}
\end{figure}

The teleoperation subspace mapping is created empirically with a relatively simple process. For each hand, the user must:
\begin{itemize}
\item Select an origin pose. Figure~\ref{origin_pose} shows the pose we chose for the human hand and the Schunk SDH robot. 
\item Determine poses which illustrate the extrema of the hand's kinematic limits along the basis vectors. It is up to the user to determine poses which 
illustrate the full range of values for each basis vector. Figure~\ref{calibration_poses} shows the poses which demonstrate these ranges for the human hand.
\item Define what the hand motions (finger spread, finger curl, and hand opening) mean in the context of the hand's kinematics, then identify which joints contribute to that motion. This is a winner-take-all approach, so a joint may only contribute to a single motion. We set joints which adduct the fingers to 1 in $\boldsymbol \alpha_H$, joints 
which open the hand to 1 in $\boldsymbol \sigma_H$, and joints which curl the fingers 
to 1 in $\boldsymbol \epsilon_H$. We then normalize the vectors to create $\boldsymbol A$. Table~\ref{tab:empirical_projection_matrix_creation} shows this process for the Schunk SDH hand.

\end{itemize}

Creating the the empirical mapping is a simple, winner-take-all, three step approach. Despite this simplicity, we show experimentally that these calculations are sufficient to meaningfully project pose space into $\boldsymbol T$ in a way that enables teleoperation for novice users.

\section{Algorithmically Defining the Subspace Mapping} \label{sec:algorithmic_mapping}

In the previous section, we rely on a human to look at the hand's kinematics, 
define each of the motions associated with the different subspace basis 
vectors, and then determine which joints contribute to that motion.

We would like to demonstrate that we can define the subspace in a 
way which is independent of hand kinematics. We also hypothesize that this 
subspace definition allows us to create a teleoperation subspace mapping for 
a hand automatically (i.e. an algorithmic mapping). 
If the algorithmic mapping can enable teleoperation for novices, this would 
demonstrate that the value of the teleoperation subspace does not derive 
exclusively from the human intuition used to create it. 

To create a subspace mapping algorithmically, we must formalize the notion 
of a hand motion in a way that does not depend on the hand's kinematics. 
We do this using objects. Hand opening can be thought of as the hand 
grasping a series of objects that grow incrementally larger. Spreading 
the fingers results from the hand grasping a series of objects whose 
curvature increases incrementally. Finger curl is binary, and can be 
defined as the difference between a precision grasp and a power grasp for 
the same object. 

\begin{figure}[t]
\centering
\vspace{0mm}
\begin{tabular}{r}
  \includegraphics[trim=0cm 0cm 0cm 0cm, clip, width=0.9\linewidth]{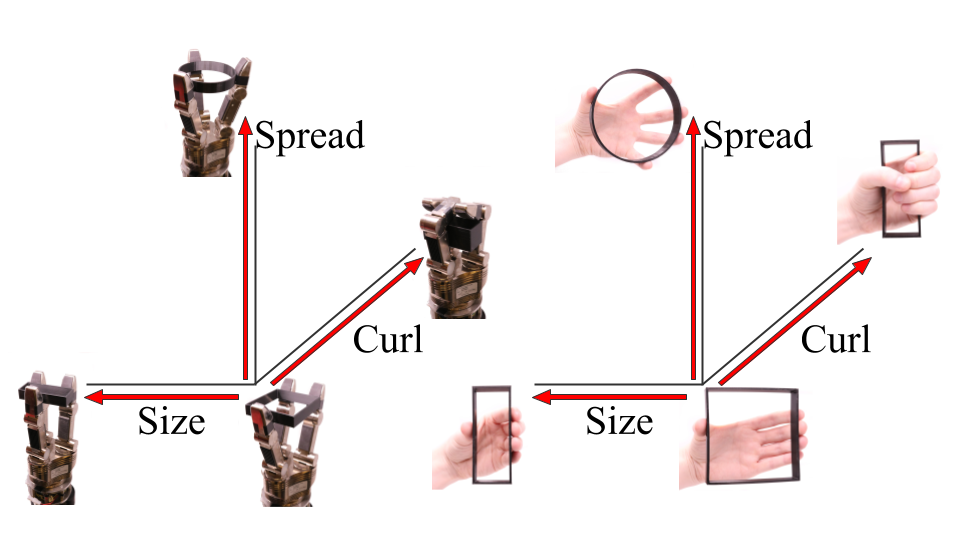}
\end{tabular}
\caption{We can formalize the hand motions that define the subspace with objects, then use this definition to predict where a grasp for that object will lie in the subspace, regardless of the hand's kinematics. This figure shows the human hand and the Schunk hand grasping the same set of objects. When an object is held by either hand, the resulting grasp will lie at the same location in the subspace $\boldsymbol T$.}
\vspace{-5mm}
\label{fig:subspace_algorithmic_paradigm}
\end{figure}

Based on this formalized notion of hand movements, we can use object 
characteristics to predict the location of a grasp in $\boldsymbol T$. 
We posit that when a hand, regardless of kinematics, is holding an 
object, we can predict where the grasp will lie in the 
teleoperation subspace, based on the object's size, shape, and the type 
of grasp used. This is illustrated in 
Figure~\ref{fig:subspace_algorithmic_paradigm}. 

If we can use the object's characteristics to predict where a grasp will lie on the subspace, we can create a set of objects which we predict will result in grasps along the basis vectors of $\boldsymbol T$. Regardless of a hand's kinematics, when the hand holds any of the objects in this set, the resulting grasp will lie along one of the basis vectors of $\boldsymbol T$. The object set we design consists of 8 objects and is described in more detail in Section~\ref{subsec:object_set}.

If a hand of any kinematic configuration grasps all the objects in our set, the result will be a set of grasps in the pose space of that hand, but that we predict can be used to find the basis vectors of $\boldsymbol T$. So, given a hand with a specific kinematic configuration, for each of the objects in our object set, we can generate a set of grasps $\mathcal{G}_{object}$. Each of the grasps $\boldsymbol g$ in $\mathcal{G}_{object}$ shows one possible way for a hand to grasp that object in a stable configuration. Each grasp $\boldsymbol g$ is an $N$ dimensional vector, where $N$ is the number of degrees of freedom for the hand. Once we have generated grasps for each of the objects, we can combine these individual sets into one grasp set $\mathcal{G}$, which encompasses all the objects:
\\[-1em]
\begin{eqnarray}
\mathcal{G} = \mathcal{G}_{object1} \cup \mathcal{G}_{object2}, ... ,\mathcal{G}_{object8}\ \nonumber\\
\mathcal{G}_{object1} = \{\boldsymbol g^{1}_{object1}, \boldsymbol g^{2}_{object1}, ... \}, \boldsymbol g \in \mathcal{R}^N  \nonumber.
\end{eqnarray}

Since $\mathcal{G}$ is a set of grasps in pose space which spans $\boldsymbol T$, we can find a model of $\boldsymbol T$ by fitting a subspace to $\mathcal{G}$. The model for $\boldsymbol T$ provides us with the subspace mapping needed to teleoperate the hand. The model of the subspace includes the origin and the directions of the basis vectors, which translate to $\boldsymbol o$ and $\boldsymbol A$ in the teleoperation mapping. We can then find $\boldsymbol \delta$ with a simple iterative method.

Once the object set has been designed, algorithmically creating a subspace teleoperation mapping requires three steps. For both the master and the slave hand, we need to:

\begin{itemize}
\item Generate a set of grasps $\mathcal{G}$ where the hand is grasping each of the objects in the object set. 
\item Fit a subspace to the grasps. The subspace model provides us with the projection between $\boldsymbol T$ and joint space. 
\item Use an iterative approach to find $\boldsymbol \delta$.
\end{itemize}

Once the mapping has been generated for a hand, it does not have to be generated again for a new master-slave pairing. For example, once we generate the human mapping, it will work with slave hand mappings generated in the same way.

We discuss the design of the object set, and the steps needed to implement teleoperation in the sections below. 

\subsection{Object Set}\label{subsec:object_set}

\begin{figure}[t]
\centering
\vspace{0mm}
\begin{tabular}{r}
    \hspace{-4mm}
    \subfloat[\label{fig:object_diagrams}]{%
      \includegraphics[trim=0cm 6cm 0cm 6cm, clip, width=1\linewidth]{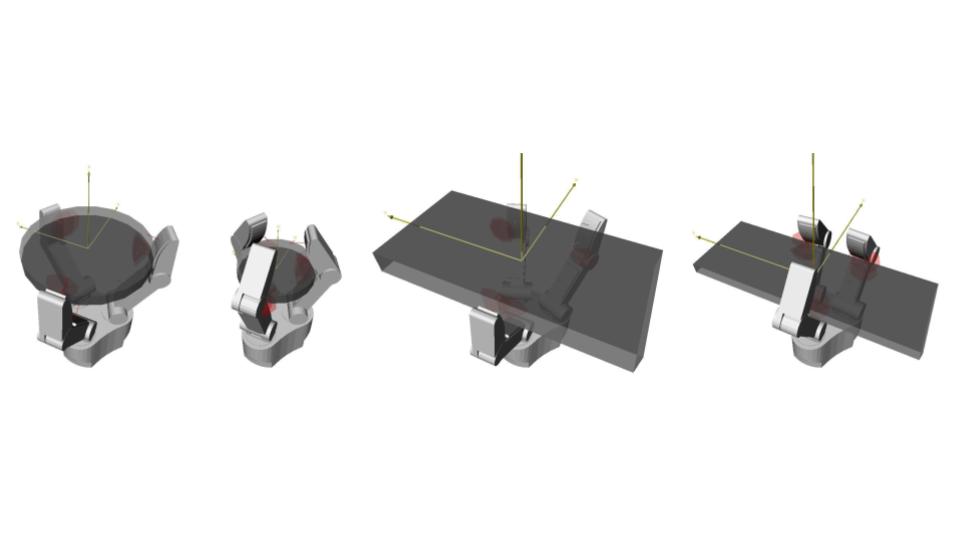}
    }
    \\
        \hspace{-7mm}
     \subfloat[\label{fig:object_locations}]{%
\includegraphics[trim=0cm 3cm 0cm 3cm, clip, width=0.9\linewidth]{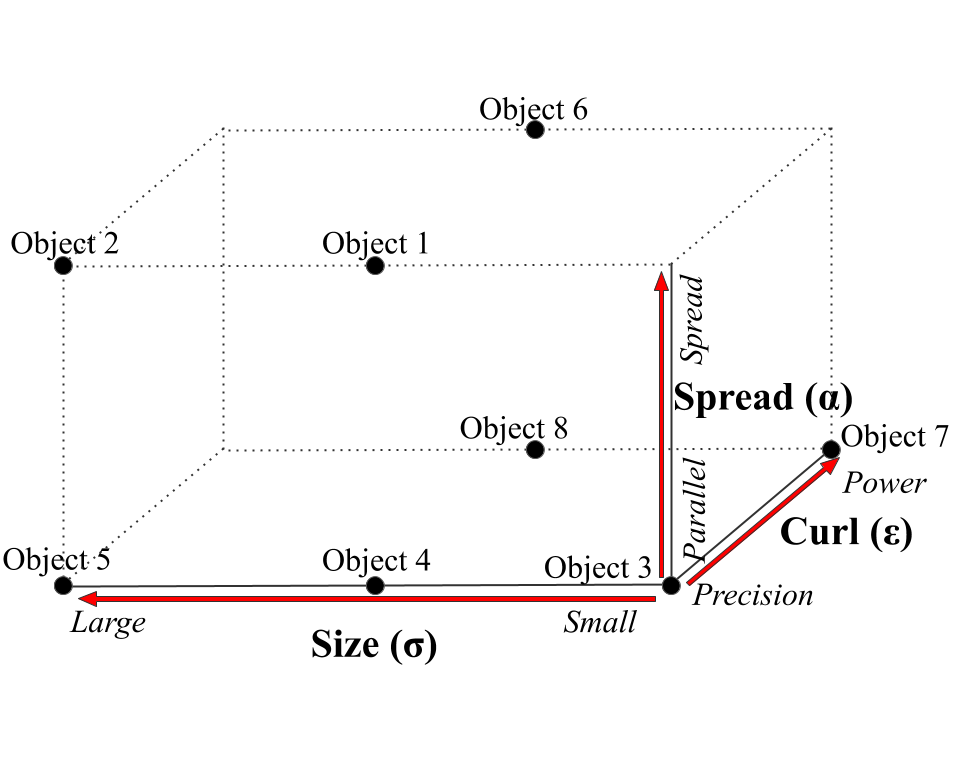}
    } 
\end{tabular}
\caption{\textbf{(a)} Four example objects from our set, held by the Schunk SDH hand. From left to right: Object 2, Object 6, Object 5, and Object 8 and \textbf{(b)} a visualization of where we predict grasps will lie in the teleoperation subspace when a hand is holding objects in the object set. This image can be used to interpret the last column of Table~\ref{tab:object_set}, where $\boldsymbol \psi = [\boldsymbol \alpha, \boldsymbol \sigma, \boldsymbol \epsilon]$.}
\vspace{-1em}
\end{figure}

We hypothesize that we can design a set of objects to elicit grasps which lie along the basis vectors of $\boldsymbol T$. Table~\ref{tab:object_set}, and Figure~\ref{fig:object_diagrams} show the objects in our set, and where in the subspace we predict hands grasping those objects will lie.

The object set consists of eight objects. We use disks and boxes as our shape primitives. We specify the type of grasp (power or precision) which must be used with each object, in order to guarantee the grasp's location along the curl basis vector of $\boldsymbol T$. In the set, there are objects that have the same dimensions, but are grasped with a different grasp type. 

The approach direction of the hand is along the z axis, and we orient the objects in the same way relative to the hand. 

We note that using a different object set would create a different subspace that would not necessarily correspond to $\boldsymbol T$. We have designed this object set specifically to fit our subspace. We selected simple objects to minimize the variance of the grasps that could be selected, both by the human and by the grasp planner. The process of object selection is driven by our intuition, but our experiments show that fitting a subspace to this object set results in a subspace which is relevant to teleoperation (though we do not guarantee that it is isomorphic to the $\boldsymbol T$ generated empirically).

\subsection{Grasp Generation}
Once we define our object set, we generate a set of grasps $\mathcal{G}$, which demonstrate how a hand of a specific kinematic configuration can hold the objects in our set in stable configurations. For robotic hands, we generate $\mathcal{G}$ using a grasp planner, and for human hands, we use human subjects.

We acknowledge that there are many ways to grasp an object. To compensate, we generate multiple grasps for each object, and use a subspace fitting method which is robust to outliers. In this way, we assume that we have sufficiently sampled grasps for the object set which would fall along the basis vectors of the subspace.

\subsubsection{Robot Datasets} \label{subsec:graspit}

To generate the robot grasps, we use a grasp planner provided by the \graspit~simulator. Given a hand and an object, the planner returns grasp configurations in which the hand stably grasps that object, ranked by the epsilon quality metric~\cite{ferrari1992}. This quality metric is a geometric method that determines the total space of possible wrenches, within certain friction constraints, for a given grasp. 

For grasp planning we apply a random search: we randomly sample an object pose (3 dimensions, we do not consider object rotation) that lies within the workspace of the hand. We also sample pre-grasp pose joint angles ($N$ dimensions, where $N$ is the number of degrees of freedom of the hand) that lie within the joint limits. We then close the fingers until they make contact with the object and evaluate the resulting grasp. In order to ensure robustness of the resulting grasps, particularly with respect to small deviations in object and pre-grasp pose, we also evaluate the grasps that arise when small perturbations are applied. Specifically we apply both positive and negative disturbances along each coordinate axis of the search space individually. Thus, for the $3+N$ dimensions from which candidate object and pre-grasp poses are sampled, we evaluate a total of $3(3+N)$ grasps. We choose the minimum quality encountered across these trials to represent the sampled grasp overall. This process is repeated until an iteration limit is reached and the sampled grasps are stored in a database.

Given a hand and an object, the planner returns up to 1,000 stable grasp configurations for that object. We parse the dataset by removing grasps which are closer than a parsing threshold $\xi$ in Euclidean distance to a higher ranking grasp. $\xi$ starts at 0.0 and is increased in intervals of 0.1. Each time $\xi$ increases, the dataset is re-parsed. This is repeated until each object has fewer than 20 grasps remaining. Therefore, the final parsed set may have a different number of grasps for each object.

The object set we present is sized to the human hand. However, some robot hands are larger than the human hand. \remindtext{3-1}{We therefore scale the objects based on hand size. The fingers of the Schunk SDH are approximately 1.5 times the size of the average human finger. So, we multiply the dimensions of the objects by 1.5 when we plan grasps for the Schunk SDH.}

\subsubsection{Human Dataset} \label{subsec:human_data}

Our grasp planner does not have a robust model of the human hand, so we generate a dataset for the human hand using grasps generated by test subjects. 

Subjects were asked to don \modifiedtext{3-9}{an} instrumented dataglove (a Cyberglove III) and grasp objects in the object set. After the subjects grasp a given object stably, their joint angles are collected from the Cyberglove. We collected grasps from five subjects. The human dataset is not parsed because there are no metrics available which would tell us how well each of the subjects grasped the objects.

\begin{table}[t!]
\vspace{3mm}
\caption{Object Set}
\label{tab:object_set}
\vspace{-3mm}
\centering
\begin{tabular}{C{.07\linewidth} C{.15\linewidth} C{.04\linewidth}  C{.04\linewidth}   C{.04\linewidth}   C{.07\linewidth}   C{.25\linewidth}}
		& & \multicolumn{3}{c}{Dimensions (in mm)}	\\
		Identifier 	& Object Primitive	& x	& y	& z	& Grasp Type & Predicted location of grasp in $\boldsymbol T$	\\ \hline

		1		& Disk			& 70	& 70	& 10	& Precision	& $\boldsymbol \psi = [1, 0.5, 0]$		 \\
		2		& Disk			& 110	& 110	& 10	& Precision	& $\boldsymbol \psi = [1, 1, 0]$	 \\
		3		& Box			& 45	& 300	& 10	& Precision	& $\boldsymbol \psi = [0, 0, 0]$	 \\
		4		& Box			& 70	& 300	& 10	& Precision	& $\boldsymbol \psi = [0, 0.5, 0]$	 \\
		5		& Box			& 100	& 300	& 10	& Precision 	& $\boldsymbol \psi = [0, 1, 0]$	\\
		6		& Disk 			& 70	& 70	& 10	& Power		& $\boldsymbol \psi = [1, 0.5, 1]$	 \\
		7		& Box			& 45	& 300	& 10	& Power		& $\boldsymbol \psi = [0, 0, 1]$ 	\\
		8		& Box 			& 70	& 300	& 10	& Power		& $\boldsymbol \psi = [0, 0.5, 1]$ 	\\

\end{tabular}
\vspace{-1em}
\end{table}

\subsection{Fitting a Subspace to a Grasp Dataset}

We hypothesized that grasps created by holding the objects in our set would exist in the pose space of the hand, but lie along the basis vectors of $\boldsymbol T$. If this is true, then we can find a model of $\boldsymbol T$ by fitting a subspace to the grasps in $\mathcal{G}$. We want the model of $\boldsymbol T$ to explain enough of $\mathcal{G}$ to enable teleoperation.

A model of $\boldsymbol T$ would provide us with the information necessary to create a teleoperation mapping for the hand. The model of the subspace consists of an origin and three $N$-dimensional orthogonal vectors, which describe the bases of the subspace. For a given hand, the origin pose of the subspace provides us with an origin pose $\boldsymbol o$ for $\boldsymbol T$, and the basis vectors provide us with a projection matrix $\boldsymbol A$.

To find the model of $\boldsymbol T$, we fit a subspace to the set of grasps $\mathcal{G}$ using \addedtext{3-6}{random sample consensus (RANSAC)}~\cite{derpanis2010
}. RANSAC is a consensus based algorithm used to find the model underlying data with a large number of outliers. The basic algorithm of RANSAC is as follows:

\begin{itemize}
\item Generate a model hypothesis using random samples from the dataset. The number of samples selected should be the minimum number needed to define your model.
\item Looking at all the points in the dataset, determine how well the hypothesis model explains/supports the data. If it is better than the best hypothesis to date, update the best model to your current hypothesis.
\end{itemize}

This process is repeated $M$ times, where $M$ is a number high enough to ensure that the probability of finding a model that is better than the current best model is sufficiently low. For our algorithm, $M = 2,000,000$. When parallelized, the runtime is 187 minutes on a computer with 24 CPUs.

Since our subspace is three dimensional, our model hypothesis consists of an origin grasp and three basis vectors. We also keep track of which of the three basis vectors corresponds to size, spread, and curl.

To generate a model hypothesis, we select random samples from the dataset. We first select an origin grasp. We specify that the origin must come from the set of grasps where the hand is holding Object 8 ($\mathcal{G}_{8}$). Preliminary tests showed the performance for this origin was the highest. We hypothesize this is because the constraints of the enveloping grasps are greater than the constraints of fingertip grasps. This gives the grasp planner (and the human) fewer options in how to grasp the objects, so the grasps are less variable.

Next, we select three additional grasps. We specify that each additional grasp must be selected from an object whose position in the subspace is identical to the origin object, except along a single basis vector. Since we have specified the origin, we randomly select one grasp from the set where the hand is holding Object 7 ($\mathcal{G}_{7}$), another grasp from $\mathcal{G}_{4}$, and the last grasp from $\mathcal{G}_{6}$. These objects correspond to the size, curl and spread directions, respectively.

After we choose four random samples, we generate our model hypothesis. We subtract the three non-origin grasps from the origin and normalize the result to find the three basis vectors of the subspace. We randomize the order of the three basis vectors, then orthogonalize these three vectors using Gram-Schmidt orthogonalization~\cite{bjorck1994}. 

We determine the quality of our model hypothesis by how many objects from the object set can be grasped using the hypothesized subspace. To determine how well the hypothesis model explains the grasp data, we find the inliers in $\mathcal{G}$ by calculating the distance from each grasp to the subspace defined by the hypothesis model. The distance $d$ from each grasp $\boldsymbol g$ to the hypothesis subspace model is found by projecting the grasp onto the subspace $\boldsymbol g_{proj}$ and then finding the distance between the true grasp and the projected grasp :
\begin{eqnarray}
\boldsymbol P = \boldsymbol \omega_1^\top \cdot \boldsymbol \omega_1 + \boldsymbol \omega_2^\top \cdot \boldsymbol \omega_2 + \boldsymbol \omega_3^\top \cdot \boldsymbol \omega_3 \\
\boldsymbol g_{proj} = \boldsymbol P \cdot (\boldsymbol g - \boldsymbol o) + \boldsymbol o \\
d = \lVert \boldsymbol g - \boldsymbol g_{proj} \rVert
\end{eqnarray}
where $\boldsymbol \omega_1, \boldsymbol \omega_2$, and $\boldsymbol \omega_3$ are the basis vectors of the hypothesis model, and $\boldsymbol o$ is the origin of the hypothesis model.
Grasps which are closer than $\xi$ (the final threshold used when we parsed the datasets) in Euclidean distance to the subspace are considered inliers:
\begin{eqnarray}
\boldsymbol g = 
	  \begin{cases}
         \text{inlier} & \text{if} \left| d \right| < \xi  \\
         \text{not an inlier} & \text{otherwise.}
      \end{cases}
\end{eqnarray}

Since we did not parse the human grasps, we simply set $\xi$ for the human dataset as 0.1.

Many RANSAC algorithms use the total number of inliers to estimate how well the model explains the data; however, we wish all parts of our subspace to fit equally well. If the grasps for a few objects contain all the inliers and grasps for all other objects are far from the subspace, we do not consider this to be a sufficiently good model, even if it has the highest total number of inliers. We want our model to be able to grasp all the objects in our dataset. So, we use a tiered metric which considers the quality of fit in all parts of the subspace.

Our tiered metric has 4 components, ranked by importance: 
\begin{enumerate}
\item Minimum number of inliers per object, over all the objects in our set. If each object has at least one inlying grasp, then we consider that model to be better than a model where one or more of the objects have no inliers, because we can grasp all the objects in our object set.
\item Number of objects which have the minimum number of inliers. If only one object has one inlier and all other objects have more than one inlier, this is preferable to all of the objects only having one inlier. 
\item Total number of inliers across all grasps. The higher the number of inliers, the better the model.
\item Sum of the distances (error) between all the grasps and the subspace. The model with the lower error is better.
\end{enumerate}

If two hypotheses tie in one or more metrics, the subsequent tier is the tiebreaker which determines the best model.

Once we have tested a sufficient number of hypotheses, the hypothesis model which explained the data the best, as defined by our metric, is considered to be the model of our subspace. 

We perform one more processing step to find our final model. \remindtext{3-5}{The same preliminary testing which indicated the best origin for the subspace model was Object 8 also showed that this was not the best origin when we combined the mappings for two hands into a complete teleoperation pipeline. For the final processing step, we choose a grasp from a different object to serve as the origin; empirically, we have found Object 1 to serve best in this role.} For the robot hand, we move the origin to the grasp from $\mathcal{G}_{object1}$ that is closest to the original subspace. For the human hand, we ask the teloperator to grasp a model of Object 1, and use the resulting pose as the subspace origin. Performing this additional step for every teleoperator also calibrates the mapping to the dimensions of their hand.

\begin{figure*}[t]
\centering
\vspace{0mm}
\begin{tabular}{C{2 cm}C{4.5cm}C{4.5cm}C{4.5cm}}
     & \underline{Human Hand} & \underline{Schunk SDH} & \underline{Two Finger Gripper} \\ 
    Empirical Mapping & 
       \includegraphics[trim=0cm 0cm 0cm 0cm, clip, width=0.9\linewidth]{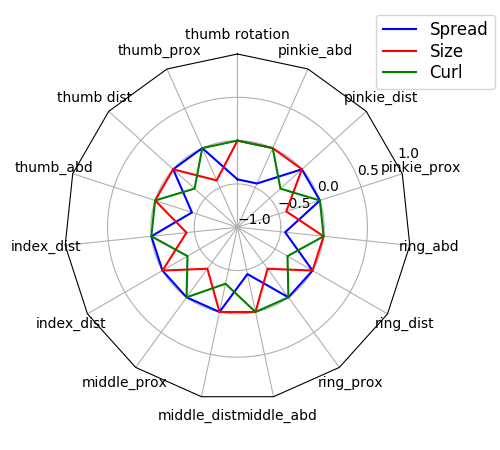} &
       \includegraphics[trim=0cm 0cm 0cm 0cm, clip, width=0.9\linewidth]{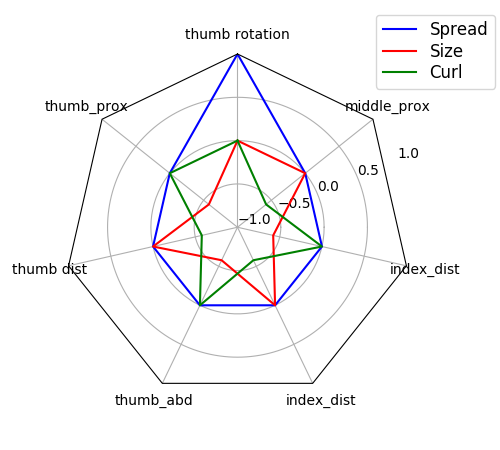} &
       \includegraphics[trim=0cm 0cm 0cm 0cm, clip, width=0.9\linewidth]{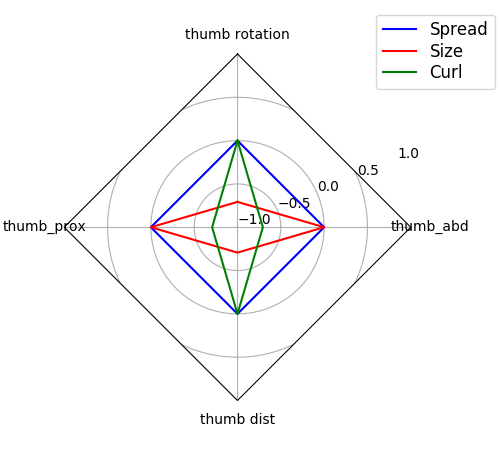} \\
    Algorithmic Mapping & 
       \includegraphics[trim=0cm 0cm 0cm 0cm, clip, width=0.9\linewidth]{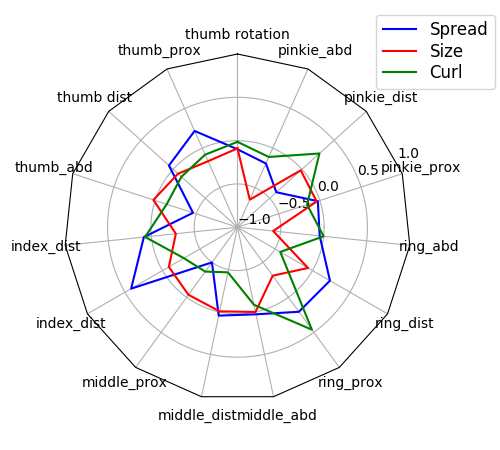} &
       \includegraphics[trim=0cm 0cm 0cm 0cm, clip, width=0.9\linewidth]{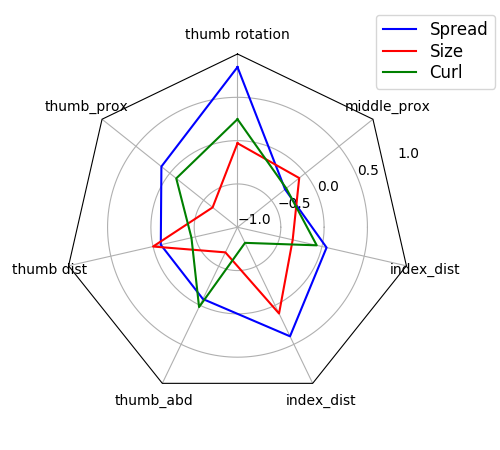} &
       \includegraphics[trim=0cm 0cm 0cm 0cm, clip, width=0.9\linewidth]{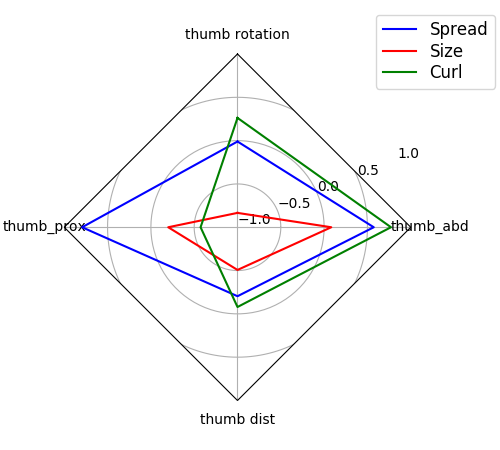}     \\
\end{tabular}
\vspace{-5mm}
\caption{Teleoperation mappings generated for the human hand, Schunk SDH, and two finger gripper, both empirically and algorithmically. Each of the spokes represents a degree of freedom for the hand, and the blue (spread), red (size) and green (curl) values along those spokes indicate the values in the $\boldsymbol \alpha_H$,  $\boldsymbol \sigma_H$, and $\boldsymbol \epsilon_H$, respectively, at that degree of freedom.}
\vspace{-5mm}
\label{fig:mappings}
\end{figure*}

\subsection{A Complete Mapping}

\subsubsection{Projection Matrix}

We use the three basis vectors of the subspace model found by RANSAC as the vectors which make up the projection matrix $\boldsymbol A$. During RANSAC, we keep track of which of the three vectors corresponds to size, spread, and curl. We use this information to determine which vector is $\boldsymbol \sigma_H$,  $\boldsymbol\alpha_H$, and $\boldsymbol\epsilon_H$, respectively. 

\subsubsection{Origin Pose}

For a robot hand, the origin of the subspace model found by RANSAC becomes $\boldsymbol o$. For a human hand, we find the origin by asking the user to perform a calibration pose at the beginning of teleoperation. A standardized pose will not work for humans because user hand size varies.

\subsubsection{Scaling Factor}

To determine the scaling factors for our mapping, we require poses for the hand at the extremes of the subspace. We could select grasps from the dataset to determine these ranges, but it is faster to use a simple assumption and an iterative solution to find them. 

For a robot, we assume that the hand will achieve its minimum and maximum value along each basis vector when the joints relevant to that basis are at some combination of their maximum and minimum values. We are given the maximum and minimum values for each joint from our robot model and the projection matrix tells us which joints are relevant to each subspace basis (if they are non-zero, they are relevant). We iterate through all the combinations of the relevant joints at their maximum and minimum values to find the set of poses which show the hand's kinematic extrema. 

For the human hand, we require the human to perform four calibration poses which will give us the ranges along each basis vector (see Figure~\ref{calibration_poses}). We require these poses because the differences in user hand size mean ranges which work for one person may not work for another. 

For both human and robot hands, we project all the poses into the subspace, using the projection matrix and the origin of our subspace model. We use the largest and the smallest value for each of the dimensions to calculate the range of that basis, and use Eq.~\ref{eq:delta} and Eq.~\ref{eq:delta_star} to find $\boldsymbol \delta$ and $\boldsymbol \delta^*$.

\subsubsection{Using the Mapping to Teleoperate}

Once we have $\boldsymbol A$, $\boldsymbol o$, $\boldsymbol \delta$, and $\boldsymbol \delta^*$ for both hands, we use Equation~\ref{eq:complete_projection_algorithm} to teleoperate.

\section{Experiments}

To validate that both the algorithmic and empirical mappings project to a subspace which is relevant to teleoperation, we asked \remindtext{2-4}{ten novice} users to complete manipulation tasks using both our mappings, and two baseline state-of-the-art mappings. \remindtext{2-4}{Five of the novices performed pick and place experiments with a Schunk SDH hand, and five performed in-hand manipulation tasks with a two fingered gripper.}

For both experiments, subjects were presented with the objects in the same order, and completed objects with one control before moving on to another control. \remindtext{3-2}{We randomized the order in which the subjects used the controls.} We did not tell subjects how the control methods worked, but gave them two minutes to play with the hand when they were introduced to a new control. The subjects gave their informed consent and the study was approved by the Columbia University IRB.

Below, we describe our mappings and our experiments. 

\subsection{Subspace Teleoperation Mappings}

We generated teleoperation mappings for the human hand, the Schunk SDH and a two fingered gripper. For each hand, we created the mappings empirically and algorithmically, using the procedures outlined in Section~\ref{sec:empirical_mapping} and Section~\ref{sec:algorithmic_mapping}. Figure~\ref{fig:mappings} shows the resulting mappings for all three hands.

\subsection{State-of-the-Art Comparisons}

We selected two state-of-the-art teleoperation mappings with which to compare our subspace mappings: 

\subsubsection{Fingertip Mapping}  We use fingertip mapping as a state-of-the-art comparison because it is one of the most common mapping methods and it is applicable to precision grasps, particularly with smaller objects~\cite{rohling1993}. To implement fingertip mapping, first, we found the cartesian positions of the thumb, index, and ring fingers of the human hand using the joint values from the Cyberglove and forward kinematics. The kinematic model for the human hand is described elsewhere~\cite{cobos2008}. We multiplied these positions by a scaling factor of 1.5, the ratio between an average human finger and the robot fingers. This ratio is 1.5 for both the Schunk SDH and the two fingered gripper. We assign each human finger a corresponding robot finger (for the two finger gripper, only the thumb and the index are used). We translated the coordinates from the hand frame into the finger frame to find the desired robotic fingertip positions.  Finally, inverse kinematics determined the joint angles which placed the robot fingertips at these positions~\cite{geng2011}.

\subsubsection{Joint Mapping} We chose joint mapping as the second state-of-the-art comparison because of its common use in the field, and because we predicted that explicit control over individual joints of the robotic fingers would be intuitive for novice users~\cite{cerulo2017}. To implement joint mapping, we assigned each of the robot joints to a corresponding human hand joint. This mapping can be found in  Table~\ref{tab:schunk_joint_mapping} for the Schunk SDH and Table~\ref{tab:seas_joint_mapping} for the two fingered gripper. The Cyberglove gave us human joint angles, and we set the corresponding joints of the robot hand to the same values.  Preliminary tests showed teleoperation is difficult if the robot thumb's proximal joint maps to the human thumb's metacarpophalangeal (MCP) joint. We therefore mapped the Schunk thumb's proximal joint and the left proximal joint of the two finger gripper to the human thumb's adductor.

We chose not to compare our mapping with a pose or synergy mapping for reasons which we enumerated in Section~\ref{sec:related_work}.

\begin{table}[t]
\vspace{3mm}
\centering
\caption{Joint mapping from the Cyberglove to the Schunk SDH}
\label{tab:schunk_joint_mapping}
\begin{tabular}{  C{.075\linewidth}  C{.325\linewidth} | C{.075\linewidth}  C{.34\linewidth} }
  \multicolumn{2}{c|}{\textbf{Cyberglove Sensor}}      & \multicolumn{2}{c}{\textbf{Robotic Hand Joints}}         \\   
  Joint Label & Name & Joint Label & Name \\ \hline
  e &   Index/Middle adduction  & 0 &   Finger 1 adduction \\   
  a &   Thumb adduction             & 1 &   Thumb proximal flexion \\ 
  b &   Thumb distal flexion    & 2 &   Thumb distal flexion \\ 

  c &   Index proximal flexion  & 3 &   Finger 1 proximal flexion\\ 
  d &   Index medial flexion    & 4 &   Finger 1 distal flexion\\ 
  f &   Middle proximal flexion     & 5 &   Finger 2 proximal flexion \\ 
  g &   Middle medial flexion   & 6 &   Finger 2 distal flexion \\ 
\end{tabular}
\begin{tabular}{r}
\hspace{-3mm}
\includegraphics[trim=5cm 3cm 5cm 2.5cm, clip,width=0.7\linewidth]{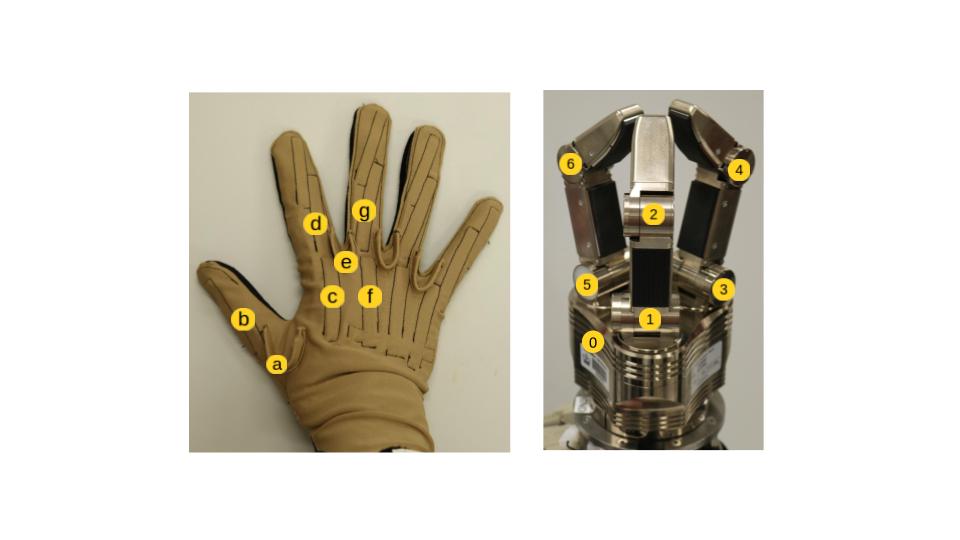}
\end{tabular}
\vspace{-8mm}
\end{table}

\begin{table}[t]
\vspace{3mm}
\centering
\caption{Joint mapping from the Cyberglove to the two finger gripper}
\label{tab:seas_joint_mapping}
\begin{tabular}{  C{.075\linewidth}  C{.325\linewidth} | C{.075\linewidth}  C{.34\linewidth} }
  \multicolumn{2}{c|}{\textbf{Cyberglove Sensor}}      & \multicolumn{2}{c}{\textbf{Robotic Hand Joints}}         \\   
  Joint Label & Name & Joint Label & Name \\ \hline
  a &   Thumb adduction  & 0 &   Finger 1 proximal flexion\\ 
  b &   Index distal flexion    & 1 &   Finger 1 distal flexion\\ 
  c &   Middle proximal flexion & 2 &   Finger 2 proximal flexion \\ 
  d &   Middle medial flexion   & 3 &   Finger 2 distal flexion \\ 
\end{tabular}
\begin{tabular}{r}
\hspace{-3mm}
\includegraphics[trim=5cm 3cm 3.5cm 2.5cm, clip,width=0.75\linewidth]{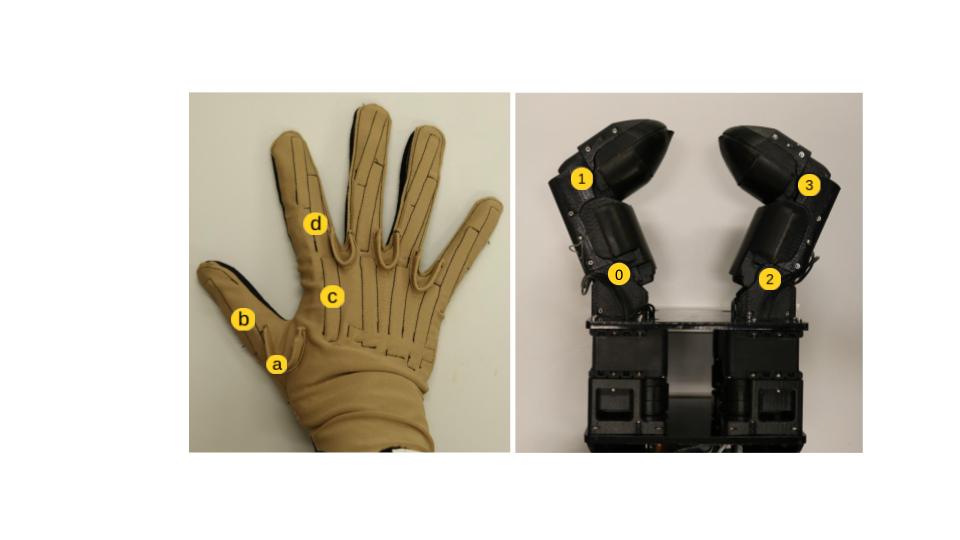}
\end{tabular}
\vspace{-5mm}
\end{table}

\subsection{Pick and Place Experiments}
We asked five novice users to complete pick and place tasks with our mappings and with state-of-the-art mappings.

\begin{figure}[t]
\centering
\vspace{2mm}
\hspace*{-1em}
\begin{tabular}{C{0.35\linewidth}C{0.62\linewidth}}
\includegraphics[trim=2.5cm 0.6cm 2.5cm 4cm, clip,width=0.7\linewidth]{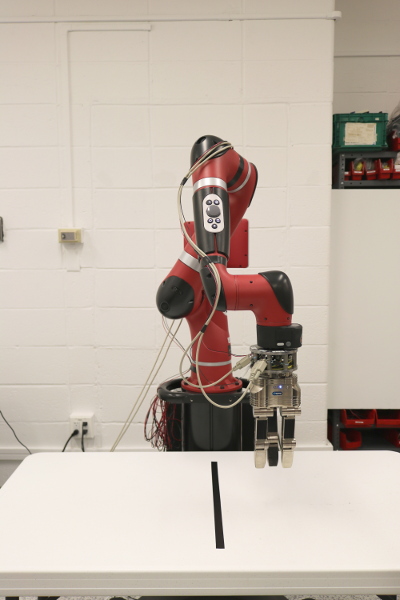} &
\includegraphics[trim=0cm 0cm 0cm 0cm, clip, width=0.7\linewidth]{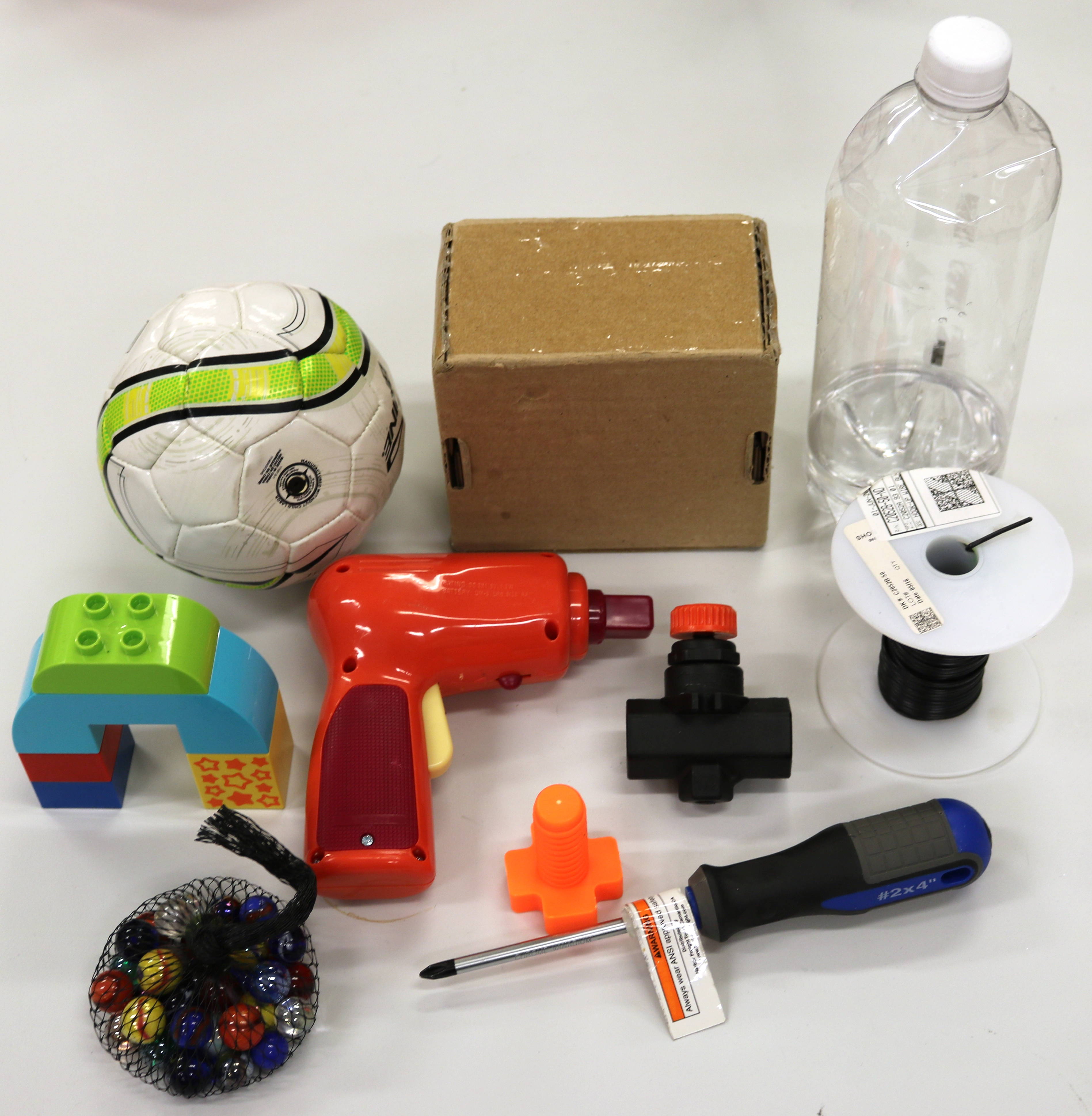} \\
\end{tabular}
\caption{(Left) Experimental set-up, and (right) object set for our pick and place experiments.}
\label{fig:pick_and_place_tests}
\vspace{-5mm}
\end{figure}

We asked our novice users to pick and place the ten objects shown in Figure~\ref{fig:pick_and_place_tests} using a Schunk SDH\addedtext{3-4}{~\cite{parlitz2013}} mounted on a Sawyer arm\addedtext{3-4}{~\cite{sawyer2020}}. The Sawyer's end effector position and orientation are controlled with a cartesian controller (completely separate from the hand control) using a magnetic tracker (Ascension 3D Guidance trakSTAR\texttrademark\addedtext{3-4}{~\cite{trakstar2020}}) placed on the back of the user's hand. \addedtext{2-5}{Using the arm, the user could move and orient the hand however they chose.} Figure~\ref{fig:pick_and_place_tests} shows the experimental setup. Subjects were asked to don a Cyberglove\addedtext{3-4}{~\cite{kessler1995}}, then pick up one object at a time and move the object across a line based on visual feedback.

\subsection{In-Hand Manipulation Experiments}

\begin{figure}[t]
\centering
\vspace{4mm}
\hspace*{-1em}
\begin{tabular}{cc}

\raisebox{0.15\height}{\includegraphics[trim=0cm 0cm 0cm 0cm, clip,width=0.35\linewidth, angle=270,origin=c]{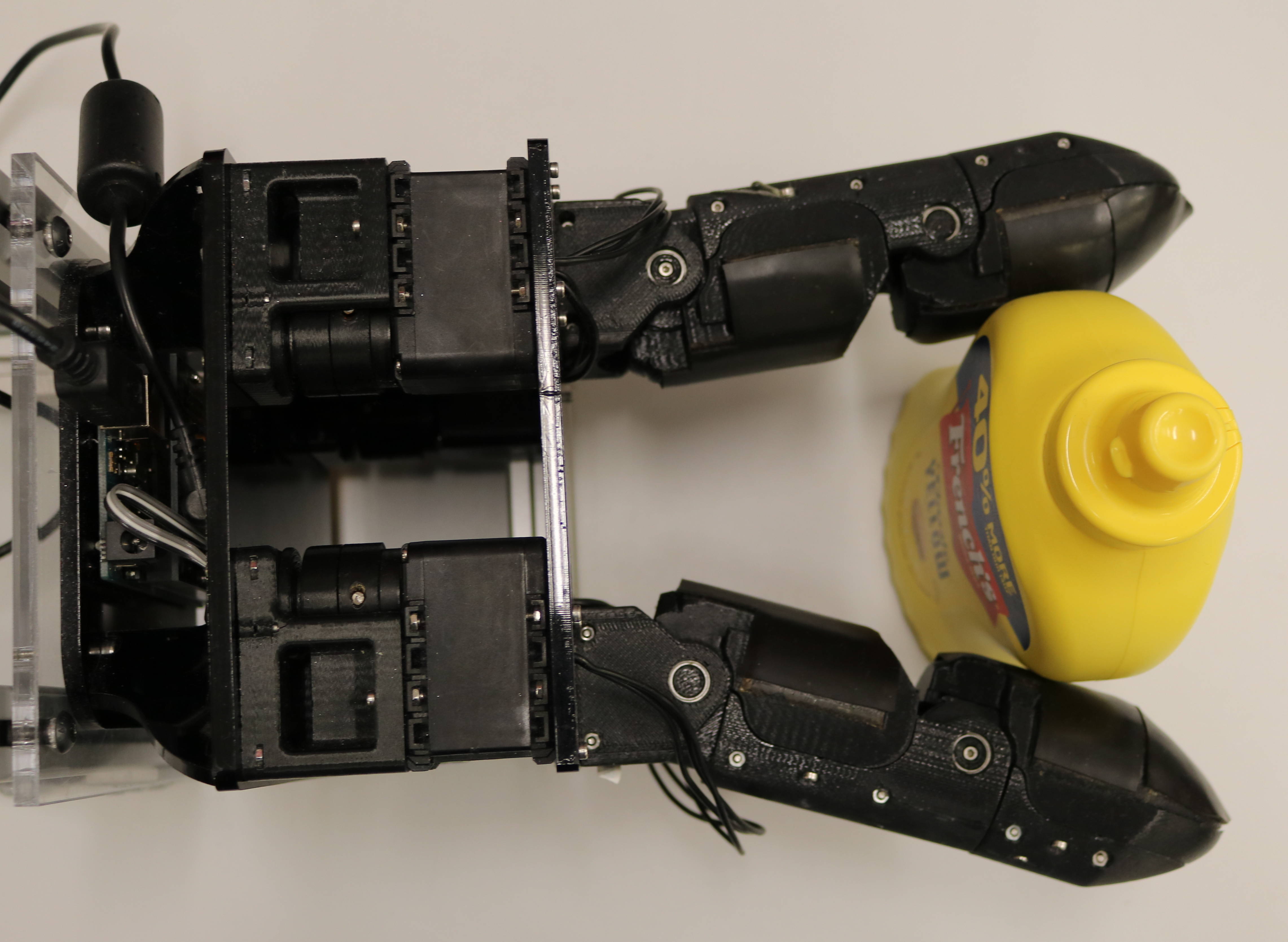}} 
\raisebox{0.15\height}{\includegraphics[trim=0cm 0cm 0cm 0cm, clip,width=0.35\linewidth, angle=270,origin=c]{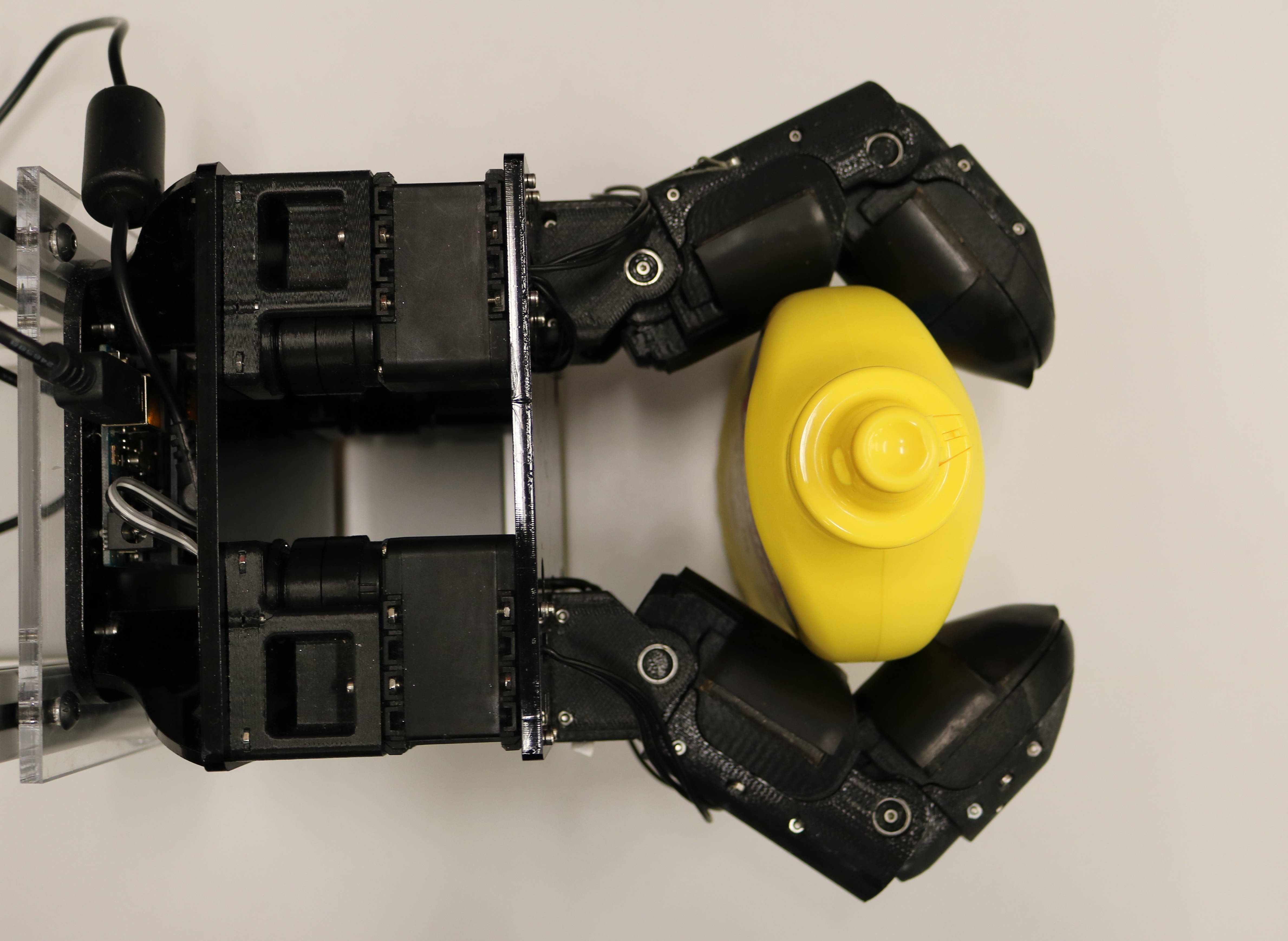}} &
\includegraphics[trim=0cm 0cm 0cm 0cm, clip, width=0.37\linewidth]{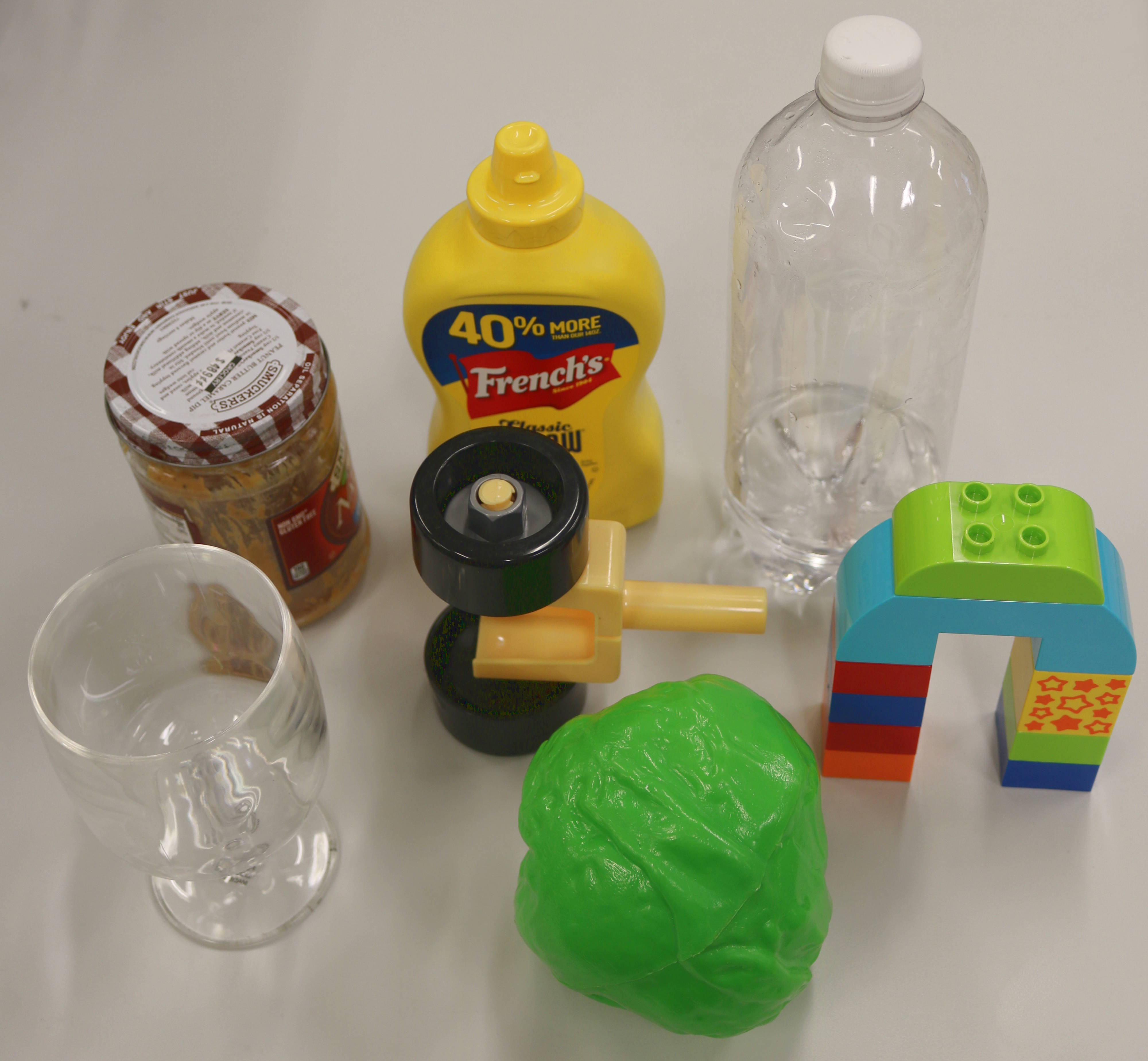} \\

\end{tabular}
\caption{(Left) Top view of experimental set-up with object in a precision grasp, (Middle) top view of experimental set-up with object in a power grasp,  and (right) object set for our in-hand manipulation experiments.}
\label{fig:in_hand_manipulation_tests}
\vspace{-5mm}
\end{figure}

We asked the other five novice users to perform in-hand manipulation tasks with a two fingered gripper\addedtext{3-4}{~\cite{chen2018}}. The gripper is stationary. An object was placed on the table between the distal links of the fingers in a precision grasp. We asked the subjects to transition the object to a power grasp by moving the object closer to the palm and enveloping it with the robot fingers. For a transition to be successful, the object had to be in contact with both the proximal and distal links on one finger and at least one link on the other finger. Figure~\ref{fig:in_hand_manipulation_tests} shows the setup and the objects used for these experiments. 

\addedtext{1-2}{
\subsection{System Latency}

To investigate the latency of our teleoperation system, we divide the problem into three parts: the first is the Cyberglove latency, which is five milliseconds (ms)~\cite{fuchs2011}. The second part is computational latency, the interval between when the Cyberglove input arrives and when the system outputs a command for the robot. For all mappings presented in this study, this latency was measured to be less than 30 ms. The third part is hardware latency, the interval from when a robot command is sent to when the robot initiates the motion. We measured this value five times and took the average. The latency is 686 ($\pm 172$) ms and 50 ($\pm 9$) ms for the Schunk and the gripper, respectively. Thus, the total latency is approximately 735 ms for the pick-and-place system (dominated by the firmware latency), and 85 ms for the in-hand manipulation system.
}

\section{Results}

For our experiments, we use three performance metrics. 
Our first metric was time to completion: how long it took for the user to perform the task. If the user did not complete the task in two
minutes for pick and place or one minute for in-hand manipulation experiments, they were considered to be unable to pick up the
object and their final time was set to the respective time limit. The in-hand manipulation time limit is shorter because there is no arm involved. 

Our second metric is the number of tries needed to complete a task. We define a ‘try’ as a completed task, an attempt where the user drops or knocks over the object, or an attempt where the user knocks an object out of the range of the robot hand. In the last two scenarios, the object is reset by the experimenter. If the subject was unable to pick the object, we report the number of tries the user took before the time elapsed. 

Our final metric is how many objects for which the task was completed: for each mapping we count how many objects for which the user was able to successfully complete the task.

\begin{table}[t!]
\vspace{1em}
  \footnotesize
  \centering
\caption{Pick and Place Experiment Results}
 \addtocounter{table}{-1}
  \subfloat[\scshape Average time to pick and place (seconds)]{%
    \hspace{-.3cm}%
\label{tab:average_time_to_pick_and_place}
\begin{tabular}{C{0.95cm}|C{1.5cm}C{1.5cm}C{1.5cm}C{1.5cm}}
 Objects    & Fingertip & Joint & Empirical & Algorithmic  \\ \hline
All & 67.8 $\pm$ 6.1 & 59.8 $\pm$ 6.2 & \textbf{25.9 $\pm$ 2.7} & 42.5 $\pm$ 4.7 \\
Small & 85.6 $\pm$ 10.9 & 92.6 $\pm$ 10.2 & \textbf{38.3 $\pm$ 7.3} &  51.3 $\pm$ 10.3 \\
Large & 49.1 $\pm$ 9.7 & 38.4 $\pm$  8.2 & \textbf{18.7 $\pm$ 2.1} & 30.0 $\pm$ 5.3 \\
Irregular & 75.1 $\pm$ 10.8 & 55.4 $\pm$ 11.2 & \textbf{22.9 $\pm$ 3.3} & 50.2 $\pm$ 9.2 \\
\end{tabular}
  }\\ \vspace{-0.1cm}
  \subfloat[\scshape Average tries to pick and place]{%
    \hspace{-.3cm}%
\label{tab:average_tries_to_pick_and_place}
\begin{tabular}{C{0.95cm}|C{1.5cm}C{1.5cm}C{1.5cm}C{1.5cm}}
Objects    & Fingertip & Joint & Empirical & Algorithmic  \\ \hline
All & 1.9 $\pm$ 0.2 & 1.9 $\pm$ 0.2 & \textbf{1.2 $\pm$ 0.1} & 1.7 $\pm$ 0.2 \\
Small & 1.9 $\pm$ 0.4 & 2.4 $\pm$ 0.4 & \textbf{1.3 $\pm$ 0.2} & 2.3 $\pm$ 0.4 \\
Large & 1.6 $\pm$ 0.3 & 1.6 $\pm$ 0.2 & \textbf{1.0 $\pm$ 0.0} & 1.4 $\pm$ 0.2 \\
Irregular & 2.5 $\pm$ 0.4 & 1.8 $\pm$ 0.4 & \textbf{1.2 $\pm$ 0.1} & 1.5 $\pm$ 0.4 \\
\end{tabular}
  }\\ \vspace{-0.1cm}
  \subfloat[\scshape Average number of objects picked]{%
    \label{tab:average_objects_picked}
    \hspace{-.3cm}%
\begin{tabular}{C{0.95cm}|C{1.5cm}C{1.5cm}C{1.5cm}C{1.5cm}}

Objects    & Fingertip & Joint & Empirical & Algorithmic  \\ \hline
All & 6.8 $\pm$ 1.0 & 7.2 $\pm$ 0.7 & \textbf{10.0 $\pm$ 0.0} & 9.0 $\pm$ 0.5 \\ 
Small & 1.4 $\pm$ 0.5 & 1.2  $\pm$ 0.5 & \textbf{3.0  $\pm$ 0.0} & 2.4 $\pm$ 0.4 \\ 
Large & 3.4 $\pm$ 0.4 &  3.8 $\pm$ 0.2 & \textbf{ 4.0 $\pm$ 0.0} &  \textbf{4.0 $\pm$ 0.0} \\
Irregular & 2.0 $\pm$ 0.5 &  2.2 $\pm$ 0.4 & \textbf{ 3.0 $\pm$ 0.0} &  2.6 $\pm$ 0.2 \\

\end{tabular}
    \hspace{.5cm}%

  }
\vspace{-0.3cm}
\end{table}

\subsection{Pick and Place Results}

We report our results as the average across all subjects. We report averages for all objects, for large objects (the box, ball, wire spool, and water bottle), for small objects (the peg, valve, and marbles), and for irregular objects (the drill, screwdriver, and lego stack). 
Irregular objects are classified as such
 because their width to length ratios and irregular shapes allow users to pick up the objects with a wide variety of grasps. Users tended to pick up the other objects with consistent grasp types. 

We report the average time to pick and place across all subjects in Table~\ref{tab:average_time_to_pick_and_place}. Across all subjects and all objects, novices using the fingertip mapping took 3 times longer than when using the empirical subspace mapping, and 1.7 times longer than when using the algorithmic subspace mapping. Similarly, joint mapping took 2.5 times longer than the empirical subspace mapping and 1.4 times longer than the algorithmic subspace mapping.

For the four combinations of objects we look at (all objects, small, large, and irregular), the empirical subspace mapping took the least amount of time, with the algorithmic subspace mapping coming in second in every case. The algorithmic subspace mapping was at least 1.6 times slower than the empirical subspace mapping for all of these object combinations. However, in turn, the state-of-the-art mappings were at least 1.3 times slower than the algorithmic subspace mapping. 

Table~\ref{tab:average_tries_to_pick_and_place} reports the average number of tries. In all cases, users were able to pick and place objects with the fewest amount of tries using the empirical subspace mapping. The algorithmic subspace mapping came in second in all cases except for small objects, where fingertip mapping was second.

Finally, we report the average number of objects the users were able to pick up with each of the mappings in Table~\ref{tab:average_objects_picked}. For all the objects, the maximum number of objects that can be picked is 10, for the small and irregular objects, the maximum is three, and for the large objects, the maximum is four.

The empirical subspace mapping allowed every novice to pick up every object. With the algorithmic subspace mapping, novices could pick up most objects, and with the state-of-the-art mappings, novices picked up the majority of objects, but still fewer than either of the subspace mapping methods.

\begin{table}[t!]
 \addtocounter{table}{1}
  \footnotesize
  \centering
  \caption{In-Hand Manipulation Experiment Results}
 \addtocounter{table}{-1}
  \subfloat[\label{tab:average_time_to_manipulate} \scshape Average time to transition from a precision grasp to a power grasp (seconds)]{%
    \hspace{-.3cm}%
\vspace{-3mm}
\begin{tabular}{C{0.95cm}|C{1.5cm}C{1.5cm}C{1.5cm}C{1.5cm}}
 Objects    & Fingertip & Joint & Empirical & Algorithmic  \\ \hline
All 	  & 16.6 $\pm$ 2.3 & 8.8 $\pm$ 1.7 & \textbf{8.5 $\pm$ 1.7} & 13.1 $\pm$ 2.0 \\
Circular  & 15.4 $\pm$ 2.6 & 7.5 $\pm$ 1.5 & \textbf{5.6 $\pm$ 0.9} &  13.6 $\pm$ 2.6 \\
Irregular & 17.5 $\pm$ 3.6 & \textbf{9.7 $\pm$ 2.8} & 10.7 $\pm$ 2.9 & 12.8 $\pm$ 2.9 \\

\end{tabular}

  }\\ \vspace{-0.1cm}
  \subfloat[\scshape Average tries to transition]{%
    \hspace{-.3cm}%
\label{tab:average_tries_to_manipulate}
\begin{tabular}{C{0.95cm}|C{1.5cm}C{1.5cm}C{1.5cm}C{1.5cm}}
Objects    & Fingertip & Joint & Empirical & Algorithmic  \\ \hline
All 	  & 1.6 $\pm$ 0.2 & 1.2 $\pm$ 0.1 & \textbf{1.1 $\pm$ 0.0} & 1.3 $\pm$ 0.1 \\
Circular  & 1.1 $\pm$ 0.1 & 1.1 $\pm$ 0.1 & \textbf{1.0 $\pm$ 0.0} & 1.1 $\pm$ 0.1 \\
Irregular & 2.1 $\pm$ 0.3 & 1.3 $\pm$ 0.2 & \textbf{1.1 $\pm$ 0.1} & 1.5 $\pm$ 0.2 \\
\end{tabular}
  }\\ \vspace{-0.1cm}
  \subfloat[\scshape Average number of objects manipulated]{%
    \label{tab:average_objects_manipulated}
    \hspace{-.3cm}%
\begin{tabular}{C{0.95cm}|C{1.5cm}C{1.5cm}C{1.5cm}C{1.5cm}}

Objects    & Fingertip & Joint & Empirical & Algorithmic  \\ \hline
All 	  & 6.6 $\pm$ 0.4 & \textbf{7.0 $\pm$ 0.0} & 6.8 $\pm$ 0.2 & \textbf{7.0 $\pm$ 0.0} \\ 
Circular  & 3.0 $\pm$ 0.0 & \textbf{4.0 $\pm$ 0.0} & 3.0 $\pm$ 0.0 & \textbf{4.0 $\pm$ 0.0} \\ 
Irregular & 3.6 $\pm$ 0.4 & \textbf{3.0 $\pm$ 0.0} & 3.8 $\pm$ 0.2 & \textbf{3.0 $\pm$ 0.0} \\

\end{tabular}
    \hspace{.5cm}%
  }
\vspace{-0.5cm}
\end{table}   

\subsection{In-Hand Manipulation Results}

We report our results as the average across all subjects. We report averages for all objects, for circular objects (the bottle, peanut butter container, and goblet), and for irregularly shaped objects (the wheels, legos, lettuce, and mustard).

We report the average time to perform the in-hand manipulation task across all subjects in Table~\ref{tab:average_time_to_manipulate}. Across all subjects and all objects, novices using the fingertip mapping took 2 times longer than when using the empirical subspace mapping, and 1.3 times longer than when using the algorithmic subspace mapping. Joint mapping performed about the same as the empirical subspace mapping and was 1.5 times faster than the algorithmic subspace mapping.

For the three combinations of objects (all, circular, and irregular), manipulation with the empirical subspace mapping took the least amount of time for all objects and the circular objects, with joint mapping taking the least amount of time for the irregular objects. In all cases, the algorithmic subspace mapping was third and fingertip mapping took the longest. 

We report the average number of tries subjects took to manipulate the objects in Table~\ref{tab:average_tries_to_manipulate}. In all object combinations, users were able to transition the objects with the fewest amount of tries using the empirical subspace mapping.

Finally, we report the average number of objects the users were able to manipulate with each of the mappings in Table~\ref{tab:average_objects_manipulated}. For all  objects, the maximum number of objects that can be manipulated is 7, for the circular objects, the maximum is three, and for the irregular objects, the maximum is four.

The joint and algorithmic subspace mappings allowed every novice to manipulate every object. For the empirical subspace mapping, one subject was not able to transition one object, and for the fingertip mapping, one subject was not able to manipulate two objects.

\section{Discussion}

We begin by discussing the teleoperation mappings generated algorithmically and empirically. In both cases, novice users were able to complete two manipulation tasks using two different non-anthropomorphic robot hands. This shows that both mappings rely on a subspace which is relevant to teleoperation and which can encompass the range of motion necessary to manipulate a variety of objects in different ways. Similarly, it shows the subspace is relevant for multiple hands.

The algorithmic subspace mapping, in particular, not only shows that the subspace we propose is relevant to teleoperation, but that the benefit of using such a subspace does not derive exclusively from the human intuition used to create the mapping. Since this mapping is created without kinematic-specific intuition from the mapping creator, and can still enable teleoperation, we conclude $\boldsymbol T$ is a concept with value even when there is no human intelligence `built into' the mapping. That being said, the empirical subspace mapping, defined with the benefit of human intuition, outperforms the algorithmic subspace mapping. Using human intuition to define the basis vectors, while not exclusively defining the value of the subspace, can make it a more powerful, intuitive control. 

We would like to emphasize that we have not designed the algorithmic mapping 
to replace the empirical mapping. In some cases, an empirical mapping can take 
significantly less time to create and also outperform the algorithmic mapping. 
The purpose of the algorithmic mapping is to show that the concept of a 
teleoperation subspace is relevant and useful for multiple hands, and that 
this relevance does not come exclusively from the human intuition built into 
the subspace via an empirical mapping. We also envision the algorithmic 
mapping to be useful for continuum robots, which do not have traditional 
finger-like structures, making the mapping difficult for a human to create, 
but we leave this to future work.

Both experiments showed that the empirical and algorithmic subspace mappings were as intuitive as or more intuitive than the state-of-the-art mappings. We measure intuitiveness as the combination of our three metrics: we hypothesize that controls which allow the user to manipulate more objects in less time, with fewer tries are more intuitive. We note that the measure of which method is preferable is a trade off between intuitiveness for the teleoperator and intuitiveness for the person who must generate the teleoperation mapping. The three metrics we have selected only measure intuitiveness for the teleoperator. 

For the pick and place experiments, the empirical subspace mapping was the most intuitive control for novices, and the algorithmic subspace mapping was the second most intuitive control. In all metrics, the empirical and algorithmic subspace mappings outperformed the state-of-the-art. The empirical subspace mapping provides the greatest advantage for small objects, but still has a significant edge for all other object combinations. The standard error we report for all the metrics is also lowest for the empirical subspace mapping. We hypothesize this means the novices were able to use the empirical subspace mapping more consistently than the other controls.

For the in-hand manipulation experiments, our empirical subspace mapping proved the most effective in terms of time to perform the experiments, followed by joint mapping and the algorithmic subspace mapping. A similar result was observed for the average number of tries required to succeed; for total objects manipulated, all three of these mappings showed similar performance, with joint mapping and algorithmic subspace mapping having a very slight advantage. This ranking is thus less definitive than for the pick and place experiments because different mappings performed better for different metrics. 

We note that the in-hand manipulation experiments lend themselves particularly well to joint mapping, which allows users to individuate the robot digits, an advantage when performing in-hand manipulation, and something which our subspace mappings do not allow. This individuation provides a particular advantage for irregular objects and is likely why joint mapping outperformed the empirical subspace mapping in time to completion for that particular object category.

These experiments show our two subspace mapping methods can generalize across different hands and different tasks.

\section{Conclusions and Future Work}

In this paper, we propose an intuitive, low dimensional mapping between the pose spaces of the human hand and non-anthropomorphic robot hands. We present an empirical algorithm to generate this mapping that leverages the user's intuition to define hand motions for a specific kinematic configuration. We also propose an algorithmic method to generate the mapping. This process is made possible by defining the subspace independently of hand kinematics, using objects to define hand motions that span the desired subspace. 

We validate both the empirical and algorithmic subspace mappings with real-time teleoperation experiments with novice users on two kinematically different robotic hands. For pick and place, our empirical subspace mapping was most intuitive for users, with the algorithmic subspace mapping performing better than state-of-the-art alternatives. For in-hand manipulation, our empirical subspace mapping performed as well as joint mapping, and better than fingertip mapping. For the in-hand manipulation experiments, the algorithmic subspace mapping was generally less intuitive for novices than joint mapping, but more intuitive than fingertip mapping. 

This is the first time, to our knowledge, that a teleoperation mapping generated without requiring a user's understanding of hand-specific kinematics has been shown to be intuitive for novices for real-time teleoperation. The fact that the algorithmic mapping can enable teleoperation shows that the subspace encodes useful information for teleoperation that does not rely exclusively on human intuition.

The future of this work could take a number of directions. Our experiments show the subspace is relevant for at least three hands, and we would like to show it is relevant for other kinematics as well. \remindtext{4-7}{It would be interesting to add more dimensions to the subspace to see if this increases dexterity while remaining intuitive for the user.} We have shown the teleoperation subspace is suitable for lower dimensional controls, like electromyography (EMG)~\cite{meeker2019} and want to validate this with more kinematic configurations. \remindtext{2-2}{We wish to show the subspace is useful for more complex tasks, like assembling machinery}\secondrevisiontext{3-1}{~\cite{roveda2020}.} \addedtext{2-1}{Finally, we wish to extend our teleoperation control for autonomous tasks or to predict user input.}

\section*{Acknowledgment}

We thank Rami J. Hamati for his hardware expertise.

\bibliographystyle{IEEEtran}
\bibliography{bib/methods,bib/related_work}

\vspace{-1.6 cm}
\begin{IEEEbiography}[{\includegraphics[width=1in,height=1.25in,clip,keepaspectratio]{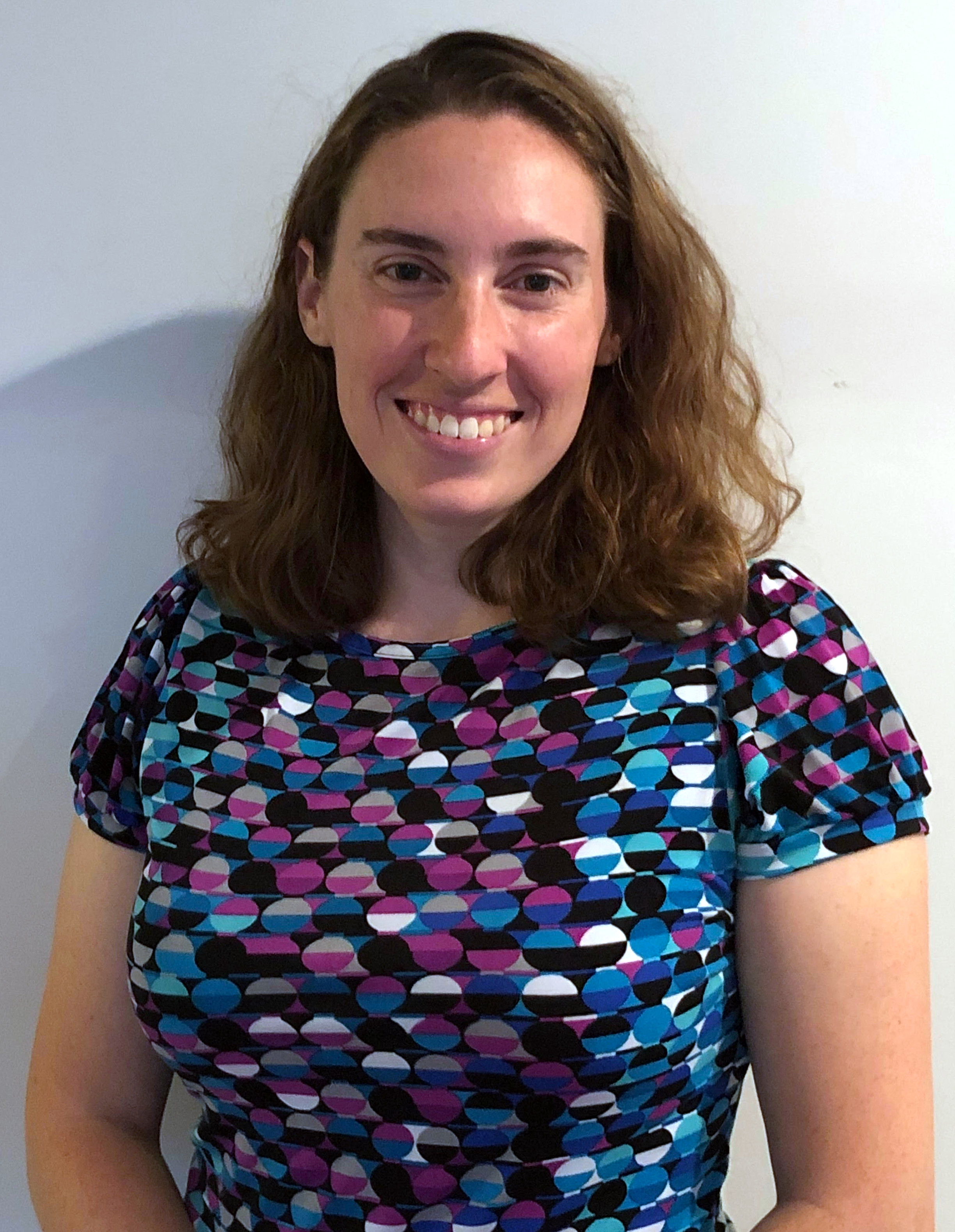}}]{Cassie Meeker}
Cassie Meeker earned a Ph.D. degree in Mechanical Engineering from the Robotic Manipulation and Mobility Lab at Columbia University in New York, NY, USA in 2020. She was a biomedical engineering research assistant at UNC Chapel Hill, NC, USA. Her research focuses on intuitive human-machine interfaces, specifically for the applications of teleoperation and rehabilitation robotics.
\end{IEEEbiography}
\vspace{-1.5 cm}
\begin{IEEEbiography}[{\includegraphics[width=1in,height=1.25in,clip,keepaspectratio]{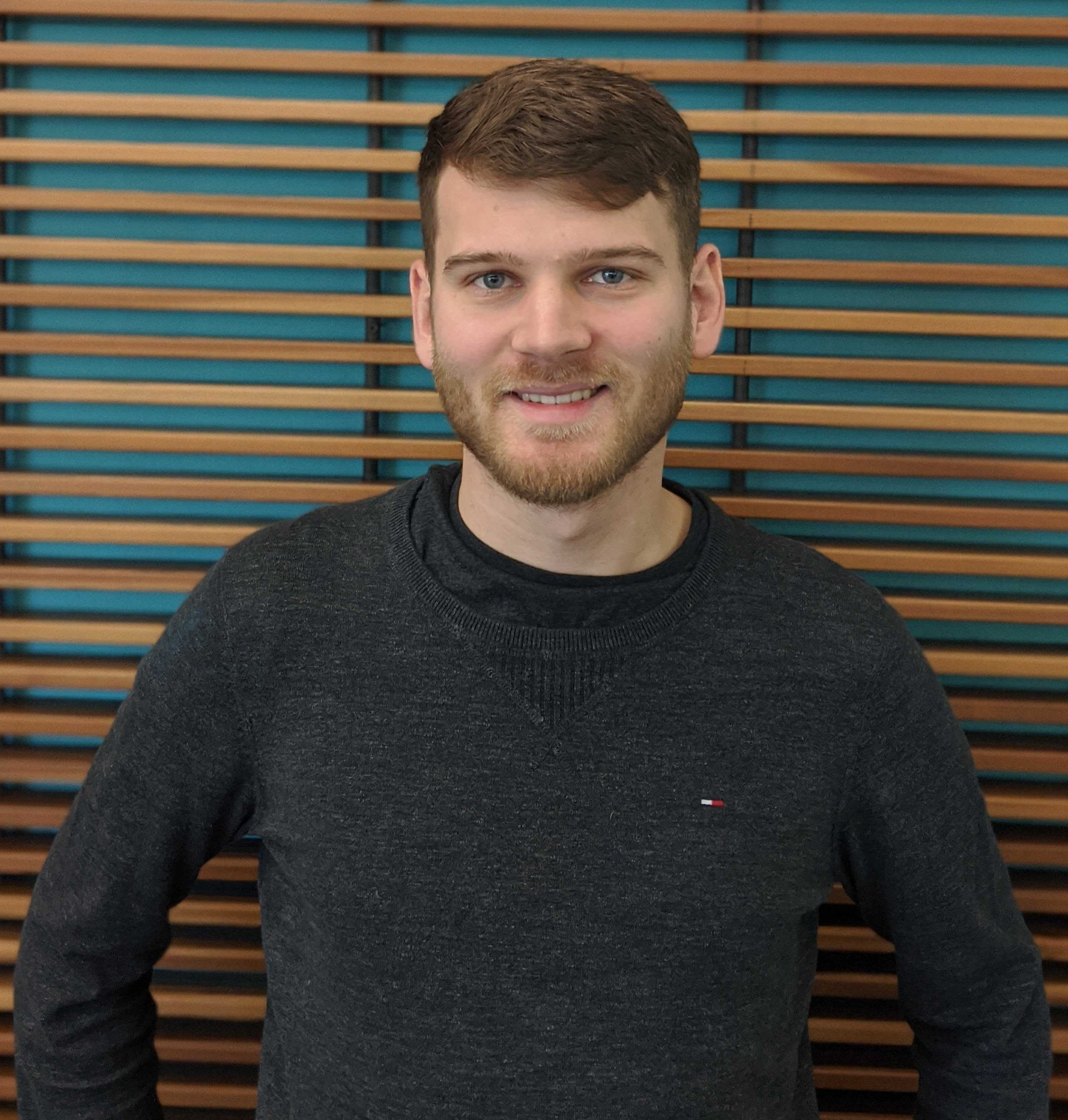}}]{Maximilian Haas-Heger}
Maximilian Haas-Heger received the MEng degree in Aeronautical Engineering from Imperial College London in 2015. Since 2015, he is a PhD candidate in the Robotic Manipulation and Mobility Lab at Columbia University in New York. His research focuses on the theoretical foundations of robotic grasping; specifically on the development of accurate grasp models and their application for dexterous manipulation.
\end{IEEEbiography}
\vspace{-1.5 cm}
\begin{IEEEbiography}[{\includegraphics[width=1in,height=1.25in,clip,keepaspectratio]{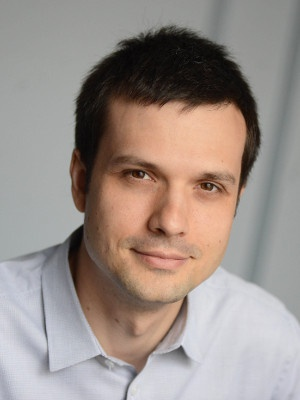}}]{Matei Ciocarlie}
Matei Ciocarlie (S'07-M'12) earned the Ph.D. degree in Computer Science from Columbia University, New York, NY, USA. He was a Research Scientist at Willow Garage, Inc., Menlo Park, CA, USA, and a Senior Research Scientist with Google, Inc., Mountain View, CA, USA. He is currently an Associate Professor of Mechanical Engineering with affiliated appointments in the Computer Science Department and Data Science Institute at Columbia University, New York, NY, USA. His work focuses on robot motor control, mechanism and sensor design, planning and learning, all aiming to demonstrate complex motor skills such as dexterous manipulation.
\end{IEEEbiography}

\end{document}